\documentclass[letterpaper]{article} 
\usepackage{aaai2026}  
\usepackage{times}  
\usepackage{helvet}  
\usepackage{courier}  
\usepackage[hyphens]{url}  
\usepackage{graphicx} 
\usepackage{natbib}  
\usepackage{caption} 
\usepackage{algorithm}
\usepackage{algorithmic}

\usepackage{threeparttable}
\usepackage{booktabs}
\usepackage{makecell}
\usepackage{amsmath}
\usepackage{multirow}
\usepackage{amsfonts}
\usepackage{siunitx} 

\usepackage{newfloat}
\usepackage{listings}
\DeclareCaptionStyle{ruled}{labelfont=normalfont,labelsep=colon,strut=off} 
\floatstyle{ruled}
\newfloat{listing}{tb}{lst}{}
\floatname{listing}{Listing}
\nocopyright

\title{FinRpt: Dataset, Evaluation System and LLM-based Multi-agent Framework for Equity Research Report Generation}

\author{
    Song Jin\textsuperscript{\rm 1}\equalcontrib,
    Shuqi Li\textsuperscript{\rm 1,2}\equalcontrib,
    Shukun Zhang\textsuperscript{\rm 3},
    Rui Yan\textsuperscript{\rm 1,4,5}\thanks{Corresponding author: rui.yan@whu.edu.cn.}
}
\affiliations{
    \textsuperscript{\rm 1}Gaoling School of Artificial Intelligence, Renmin University of China, Beijing, China\\
    \textsuperscript{\rm 2}King Abdullah University of Science and Technology, Thuwal, Saudi Arabia\\
    \textsuperscript{\rm 3}School of Smart Governance, Renmin University of China, Suzhou, China\\
    \textsuperscript{\rm 4}School of Artificial Intelligence, Wuhan University, Wuhan, China\\
    \textsuperscript{\rm 5}School of Computer Science and Artificial Intelligence, Wuhan Textile University, Wuhan, China\\
    jinsong8@ruc.edu.cn, shuqi.li@kaust.edu.sa,
    zhangshukun@ruc.edu.cn, rui.yan@whu.edu.cn
}

\usepackage{bibentry}

\begin{document}

\maketitle

\begin{abstract}
While LLMs have shown great success in financial tasks like stock prediction and question answering, their application in fully automating Equity Research Report generation remains uncharted territory. In this paper, we formulate the Equity Research Report (ERR) Generation task for the first time. To address the data scarcity and the evaluation metrics absence, we present an open-source evaluation benchmark for ERR generation - FinRpt. We frame a Dataset Construction Pipeline that integrates 7 financial data types and produces a high-quality ERR dataset automatically, which could be used for model training and evaluation. We also introduce a comprehensive evaluation system including 11 metrics to assess the generated ERRs. Moreover, we propose a multi-agent framework specifically tailored to address this task, named FinRpt-Gen, and train several LLM-based agents on the proposed datasets using Supervised Fine-Tuning and Reinforcement Learning. Experimental results indicate the data quality and metrics effectiveness of the benchmark FinRpt and the strong performance of FinRpt-Gen, showcasing their potential to drive innovation in the ERR generation field.
All code and datasets are publicly available. 
\end{abstract}


\section{Introduction}

Recently, Large Language Models (LLMs) have reshaped the field of natural language processing and presented remarkable capabilities in many specialized domains across medicine~\cite{tan2024medchatzh}, law~\cite{izzidien2024llm}, physics~\cite{polverini2024understanding}, and finance~\cite{wu2023bloomberggpt, xie2023pixiu}, etc. Within the financial domain, many recent studies have shown great progress in leveraging advanced LLMs into some traditional financial tasks, such as sentiment analysis~\cite{zhang2023sentiment}, information extraction~\cite{sharma2023financial,hamad2024fire}, question answering~\cite{yang2023fingpt}, etc., which typically focus on short summaries or brief descriptions. 

The media produces a large volume of information about individual companies every day, which includes both considerable noise and valuable insights. Equity research reports (ERRs)~\cite{siantar2024implementation} play a crucial role in filtering and summarizing this information, providing investors with an in-depth assessment of the company's financial state, market position, and investment potential.
An ERR usually consists of many segments, such as a company's financial status statement, risk evaluation, stock trend prediction, etc. Writing it requires specialized financial knowledge and experience, as well as a deep understanding of financial markets, industry trends, and company development, and is usually done by professional analysts. 

Recent advancements in LLMs~\cite{yang2023fingpt,wu2023bloomberggpt}, have made automated ERR generation~\cite{jejeniwa2024comprehensive,adeyeri2024automating} a feasible endeavor, which could shorten the time required for company-related information collection and analysis, providing timely insights and recommendations to organizations and researchers, allowing them to respond more quickly to market changes and industry trends. 
Additionally, ERRs not only provide comprehensive company analyses but also offer good explanations for stock trends, paving the way for improved stock price prediction and advancing other Fintech applications. However, generating ERRs automatically remains an unexplored area due to the following reasons.

From the benchmark perspective, data scarcity ~\cite{yagamurthy2023natural} is a major obstacle. The input information is typically unstructured and scattered across multiple sources, such as company announcements, industry reports, historical stock prices, and news articles, making it difficult to integrate these diverse data types into a cohesive and standardized format and further map it to the final ERR. Most existing evaluation metrics for generative financial tasks focus primarily on assessing the capabilities of the methods from an NLP view, such as ROUGE-L, and BERTScore~\cite{xie2023pixiu}, which are insufficient. Firstly, the evaluation framework should include metrics that assess the accuracy of key indicators in the generated report, such as the cash flow. Secondly, evaluating the accuracy of stock trend predictions is crucial, as it directly influences the potential gains and losses for investors. Finally, since ERRs rely heavily on finance knowledge, evaluating the professionalism of the generated reports is another essential component of the evaluation system.

On the method side, relying only on a single LLM~\cite{wu2023bloomberggpt, yang2023fingpt} to generate such a complex financial report is hard to achieve. Recently, many LLM-based financial multi-agent frameworks have been developed that could deal with sophisticated scenarios, such as FinAgent~\cite{zhang2024finagent} for trading and FinMem~\cite{yu2024finmem} for decision-making. An ERR typically consists of multiple sections, each of which needs to reflect an aspect of a company and follow a coherent and logical structure. To tailor a multi-agent framework for generating ERRs is also an issue that requires attention.

To bridge the aforementioned gap, this work makes the first attempt to face the ERR generation task directly and address it. Code\footnote{https://github.com/jinsong8/FinRpt} and datasets\footnote{https://huggingface.co/datasets/jinsong8/FinRpt} are public. The main contributions of our work could be summarized as follows:

\begin{itemize}
    \item We formally define the task of ERR generation for the first time, which could be generalized to report generation in other domains.

    \item We establish an open-source ERR generation evaluation benchmark, named FinRpt, consisting of a Chinese-English ERR dataset and a comprehensive evaluation system. The Dataset Construction Pipeline can automatically generate high-quality ERR data, which could be generalized to the dataset construction in other similar tasks.
    
    \item To tackle the newly defined ERR generation task, we tailor a multi-agent framework called FinRpt-Gen, which decomposes the complex task and assigns nine agents to address it collaboratively. Furthermore, we train these agents using Supervised Fine-Tuning (SFT) and Reinforcement Learning (RL), enabling them to achieve optimal performance.
   
\end{itemize}

\section{Related Work}

\noindent\textbf{Financial Agents} 
Many LLM-based financial agents are developed to tackle complex tasks. FinMem~\cite{yu2024finmem} is designed for automated trading. It is optimized by fine-tuning the agent’s perception range and character profiles, which improves trading performance and results in higher cumulative investment returns. ~\citet{zhang2024finagent} features a dual-level reflection module and a diversified memory retrieval system, which enhance its ability to make trading decisions. FinCon~\cite{yu2024fincon} is structured with a hierarchical manager-analyst model, drawing inspiration from real-world investment firms. It also features a dual-level risk control system to optimize investment strategies and effectively mitigate risks. CryptoTrade~\cite{li2024reflective} integrates both on-chain and off-chain data sources and employs a reflective mechanism to enhance trading strategies.

\noindent\textbf{Evaluation of Financial LLMs}
There are many evaluation benchmarks for the financial domain, such as 
MME-Finacne~\cite{gan2024mme},  FinanceBench~\cite{islam2023financebench}, BizBench~\cite{koncel2023bizbench},  FinMME~\cite{luo2025finmme}, PIXIU~\cite{xie2023pixiu} and FinBen~\cite{xie2024finben}, which mainly focus on Financial NLP capabilities like information extraction and question answering. Among these, finance-related multiple-choice questions~\cite{xie2023pixiu} are used to assess the model's understanding of financial knowledge. Common metrics like ROUGE~\cite{lin2004rouge} and BERTScore~\cite{zhang2019bertscore} are widely used to evaluate alignment, factual consistency, and information retention.
However, this focus limits the ability to comprehensively assess LLMs across a broader range of complex financial applications.
Equity Research Report generation encompasses many of these evaluation demands, making it crucial to design a specialized benchmark that can effectively assess ERR generation capabilities and foster its advancement.

\noindent\textbf{Equity Research Reports Generation}
Some existing multi-agent frameworks are tailored for similar tasks with ERR generation. FinRobot ~\cite{yang2024finrobot} is an open-source AI agent platform for financial applications. By prompting LLMs directly, the system could generate well-structured financial analysis.
A key limitation of FinRobot is its reliance on fixed annual reports, which restricts its ability to use real-time or diverse data.
Another notable example is FinReport~\cite{li2024finreport}, an explainable stock earnings forecasting model. 
Unlike ERR, the report generated by the model emphasizes explanations of stock predictions and risk assessments obtained through specially trained modules. 
~\citet{fons2025ai} focuses on using LLMs to generate analytical reports for financial time series. However, due to the complexity of ERR generation task, there is still a need to design a customized framework specifically tailored to address it effectively.

\begin{figure*}
 \centering
\includegraphics[width=0.95\linewidth]{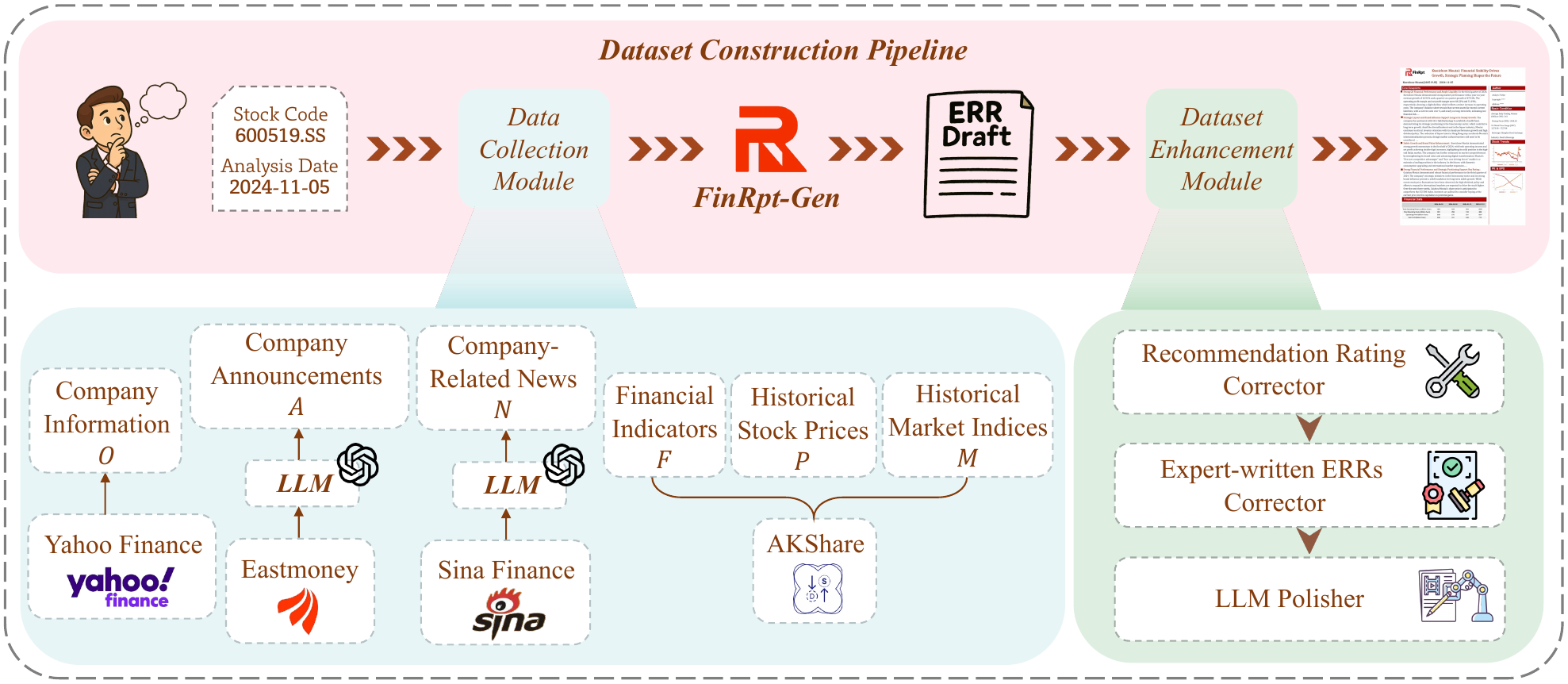}
\caption{The Dataset Construction Pipeline, Data Collection Module, and the Dataset Enhancement Module.}
\label{fig:data_collection_construction}
\end{figure*}

\section{FinRpt: Task, Benchmark and Method}

\subsection{ERR Generation Task Formulation}
\label{sec:task}
This work formally defines the ERR generation task for the first time. Given a company's stock ticker $s$ and the research date $t$, the system automatically gathers and structures recently relevant information, and then utilizes it to generate an ERR $R$. This setup replicates the workflow of a real-world research analyst when drafting an ERR.

In this paper, the input information source $S=[O, F, A, N, P, M]$ includes Company Information $O$, Financial Indicators $F$, Company Announcements $A$, Company-related News $N$, Historical Stock Prices $P$, and Historical Market Indices $M$. To define the output ERR format, we summarize that an ideal ERR of a company should at least include 6 key segments, despite varying formats across securities firms: Financial Analysis $R_{fin}$, News Analysis $R_{news}$, Management and Development Analysis $R_{manage}$, Risks Analysis $R_{risk}$, Investment Potential Assessment $R_{invest}$, and Recommendation Rating $R_{rec}$ (recommend a buy rating or a sell rating). We show a generated ERR case in Figure~\ref{fig:en_report} in the Appendix.

\subsection{Dataset}

To address the issue of data scarcity for this task, we construct a high-quality ERR generation dataset. In this section, we will thoroughly describe both the Data Collection Module and the Dataset Construction Pipeline. To enhance clarity, we have visualized the entire process in Figure~\ref{fig:data_collection_construction}, which illustrates the sequential steps involved in both data collection, dataset construction, and dataset enhancement.

\subsubsection{Data Collection Module}

High-quality data ensures that the analysis, predictions, and insights derived from it are meaningful and trustworthy. Thus, a well-designed data collection module is necessary, which should be capable of gathering key information from various credible sources that provide timely and comprehensive information about a company. 

Building on insights from previous research~\cite{zhang2024finagent, penman2013financial, greenwald2020value, yu2024finmem, zhang2024ai, yu2024fincon, fatemi2024finvision}, we carefully selected six valuable and complementary types of company-related data and integrated them into our data collection module:

(1) Company Information $O$: providing foundational company information~\cite{yu2024fincon}.
(2) Financial Indicators $F$: reflecting the company's operational status and released quarterly~\cite{fatemi2024finvision}.
(3) Company Announcements $A$: about significant changes in investment decisions, personnel changes, or unexpected events, reflecting the company's management and development~\cite{zhang2024ai}. Besides, we utilize GPT-4o-mini to summarize the announcements.
(4)Company-related News $N$: reflecting events related to a certain company, influencing investor sentiment, and impacting stock trends~\cite{yu2024finmem}. Similar to announcements summarization, GPT-4o-mini is used to summarize news content and filter out news irrelevant to the company's stock. Besides, we remove brief articles and leverage BERTScore~\cite{zhang2019bertscore} and MinHash~\cite{broder1997resemblance} to de-duplicate similar news.
(5) Historical Stock Prices $P$: reflecting the value of a company to some extent and providing valuable insights into the potential investment assessment~\cite{zhang2024finagent}.
(6) History Market Indices $M$: reflecting market conditions, as well as investor enthusiasm and confidence~\cite{yoo2021accurate}.


\subsubsection{Dataset Construction Pipeline}
\label{sec:data_construction}

To bridge the gap of data scarcity for ERR generation, we construct an ERR dataset, which consists of 800 stocks in the CSI800 Index of the Chinese market. The corresponding companies generally have a high market value, which results in a substantial amount of information being available in the media. The data range is from September 3, 2024, to November 5, 2024, with intervals of one week between analysis dates, resulting in 10 analysis dates for each company stock. The dataset has 6825 data samples (each sample including the input source information and the corresponding ERR).
First, we use the Data Collection Module to gather the input information $S=[O, F, A, N, P, M]$ for each stock ticker $s$ and analysis date $t$ forming an input $(s, t, S)$. We then apply a filtering process to enhance the quality of input data $(s, t, S)$ that excludes data lacking financial indicators $F$, those with fewer than two news articles $N$, and those with summarized announcement $A$ lengths under 300 Chinese characters.

Then we apply the multi-agent framework FinRpt-gen, introduced in the next Section, with GPT-4o as each LLM agent, to generate ERRs $R$ automatically, which leads to a complete data input-output sample $(s, t, S, R)$. To align the generated ERRs with the expert-written ERRs, we develop a Dataset Enhancement Module to enhance the quality of generated ERRs: 

(1) Recommendation Rating Corrector: for each sample $(s, t, S, R)$, the recommendation rate $R_{rec}$ in $R$ is compared to the ground truth trend label. If they are not consistent, the ERR is regarded as invalid, and this sample will be re-inferred until the correct predictions are generated. (2) Expert-written ERRs Corrector: for each sample $(s, t, S, R)$, we retrieve reports related to the stock $s$ during the week preceding the analysis date $t$ from Eastmoney as the reliable ERRs $R_{experts}$. Then the retrieved reports $R_{experts}$ along with the generated reports $R$, are used to prompt an LLM GPT-4o to review and refine the information accuracy, logical consistency, and writing style. The detailed prompt is shown in Table~\ref{tab:corrector_prompt} in the Appendix.
(3) LLM Polisher: last, we input each ERR $R$ into an LLM GPT-4o for writing polish, enhancing its readability, coherence, and logical flow. 

Based on the above processing steps, the high-quality ERR dataset FinRpt is constructed, including 6,825 ERRs. We also provide a corresponding English-translated version. It could be used for ERR generation evaluation, supervised fine-tuning, and reinforcement learning.

\subsubsection{Dataset Statistics}

\begin{figure}[h]
 \centering
\includegraphics[width=\linewidth]{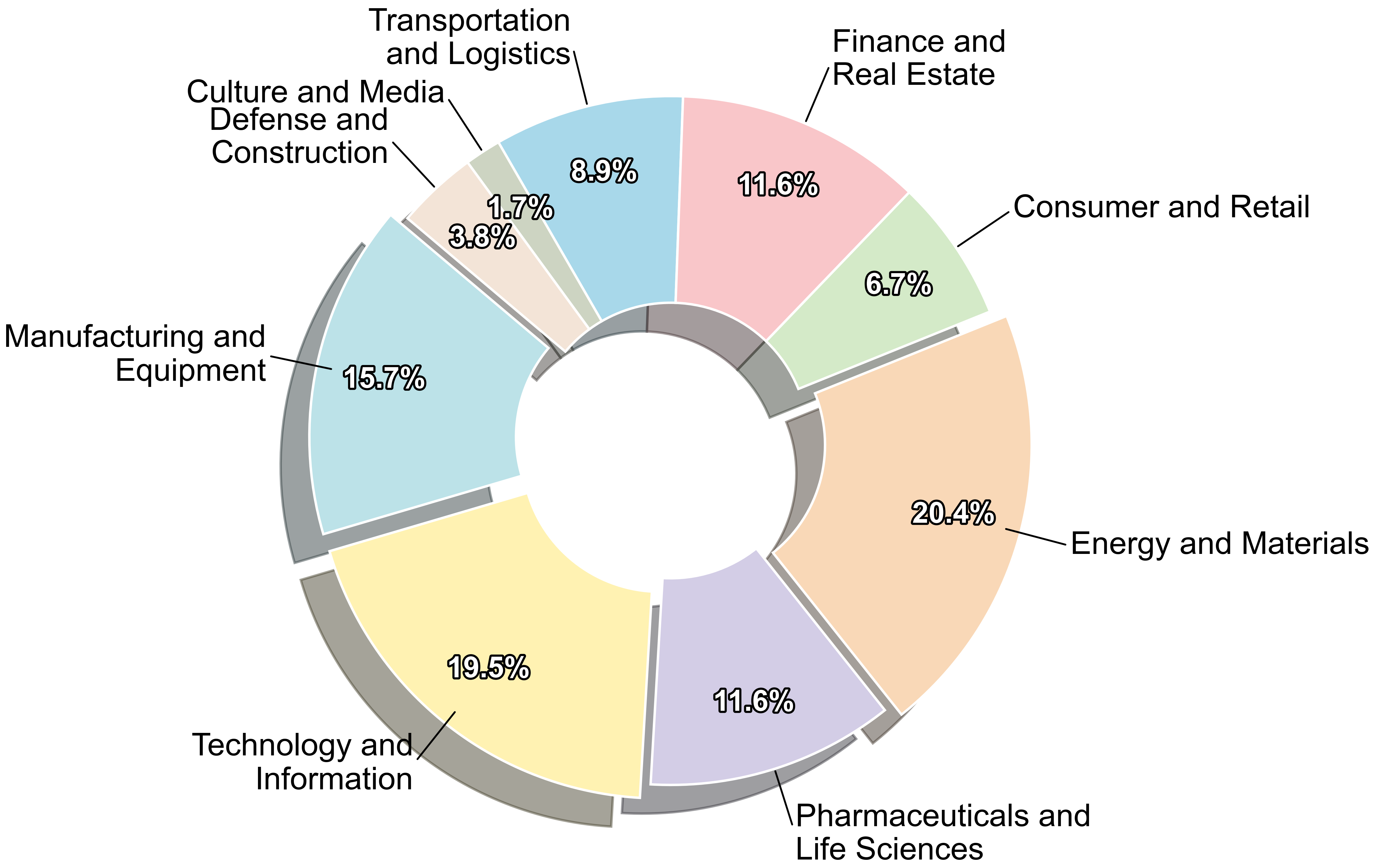}

\caption{The proportion of reports from different industries of the FinRpt dataset.}
\label{fig:industry_pie}
\end{figure}

We analyze several statistics of the constructed dataset - FinRpt. It contains a total of 6,825 reports from 2024-09-03 to 2024-11-05. On average, there are about 9 reports per stock, with a total of 683 reports per analysis date. The detailed industry-wise statistics are shown in Figure~\ref{fig:industry_pie}. 

We partitioned the dataset as follows: data before 2024-10-31 was randomly split into a training set and a validation set with a 9:1 ratio. Samples after 2024-10-31 were used as the test set. As a result, the training set contains 5,556 samples, the validation set contains 617 samples, and the test set consists of 652 samples.

\subsection{Proposed Baseline: FinRpt-Gen}
\label{sec:finrpt-gen}
\subsubsection{FinRpt-Gen}

\begin{figure}
 \centering
\includegraphics[width=0.9\linewidth]{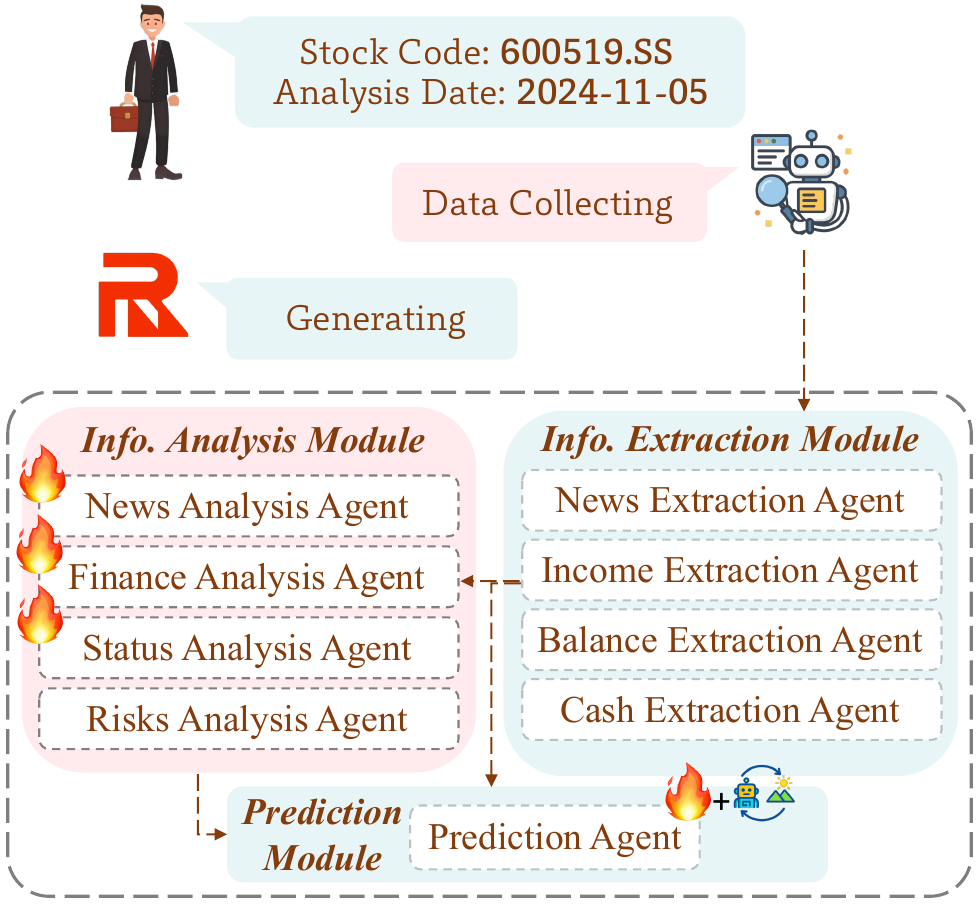}
\caption{The framework of the proposed FinRpt-Gen.}
\label{fig:framework}
\end{figure}

The task of ERR generation requires the model to have extensive financial knowledge, a standardized report writing style, and exceptional logical analysis and forecasting abilities. In this work, we propose FinRpt-Gen, as shown in Figure~\ref{fig:framework}, which is the first multi-agent framework specifically designed for the ERR generation task. Given the constructed dataset FinRpt, FinRpt-Gen consists of three modules: an Information Extraction Module, an Analysis Module, and a Prediction Module, involving nine agents playing different roles. We show the prompt examples for every agent in the Appendix.

\noindent\textbf{Information Extraction Module}
The information extraction module extracts related information from the given input data $(s, t, S)$. This module involves four different agents: 

(1) News Extraction Agent: ranking the provided news articles $N$ by the impact of the news on the stock $s$ and outputting the top 10 news articles most likely to influence stock prices. 
(2) Income Extraction Agent: given the income statement in financial indicators $F$, extracting key financial metrics such as revenue, net income, earnings per share, etc.
(3) Balance Extraction Agent: given the balance sheet in financial indicators $F$, focusing on key financial indicators such as assets, liabilities, and equity.
(4) Cash Extraction Agent: 
given the cash flow statement in financial indicators $F$, focusing on cash from operations, investing, and financing activities.

Based on the well-designed prompts and the table understanding ability of LLMs~\cite{sui2024table}, these agents can effectively extract key financial indicators and news for further analysis.
According to the ERR format that has been defined previously, we devise the following Analysis Module and Prediction Module to systematically complete the six specific sections of an ERR $R = [R_{fin}, R_{news}, R_{manage}, R_{risk}, R_{invest}, R_{rec}]$.

\noindent\textbf{Information Analysis Module}
(1) Finance Analysis Agent: given the output of the Income, Balance, Cash Analyst Agent, this agent summarizes the company's financial health, profitability, and cash flow position, generating the financial analysis $R_{fin}$.
(2) News Analysis Agent: given the output of the News Analyst Agent, this agent emphasizes how the selected news may affect the future stock performance, then produces news analysis $R_{news}$.
(3) Status Analysis Agent: derives the management and development
analysis $R_{manage}$ from recent company announcements.
(4) Risk Analysis Agent: integrates the financial, news, management, and development analysis content from the aforementioned analysis agents to analyze the key risks $R_{risk}$ that should be paid attention to.

\noindent\textbf{Prediction Module}
The Prediction Agent in the prediction module collects the analysis content of $R_{fin}$, $R_{news}$, $R_{manage}$ and $R_{risk}$ along with the historical stock prices $P$ and history market indices $M$, then forecasts the investment potential assessment $R_{invest}$ and the recommendation rating $R_{rec}$. A recommendation rating refers to an evaluation or assessment given to an investment that reflects the analyst’s opinion or suggestion regarding its performance potential, which is typically categorized as ``buy'' or ``sell''.

\subsubsection{Supervised Fine-Tuning (SFT)}

Within the FinRpt-gen framework, we focus on fine-tuning the four most critical agents: Finance Analysis Agent, News Analysis Agent, Status Analysis Agent, and Prediction Agent. These agents handle the most complex tasks of generating deep, professional insights. We use demonstration samples from the corresponding sections of the FinRpt dataset. For example, given a data sample $(s, t, S, R)$, for the Finance Analysis Agent, the input is the content generated from the Income Extraction Agent, Balance Extraction Agent, and Cash Extraction Agent. The output is the Financial Analysis $R_{fin}$ section in $R$. The fine-tuning leverages SFT with LoRA~\cite{hu2022lora}, aiming to learn a set of low-rank adapter parameters $\Delta\theta$ to maximize the likelihood of generating the target text $Y$ given an input $X$. The optimization objective is formulated as:
$$
\max_{\Delta\theta} \sum_{(X, Y) \in D_{\text{demo}}} \log P(Y | X; \theta_0 + \Delta\theta),
$$
where $D_{\text{demo}}$ is the demonstration dataset for the respective agent, $\theta_0$ represents the original parameters of the pre-trained model, and $\Delta\theta$ are the low-rank update parameters learned via LoRA.

\subsubsection{Reinforcement Learning (RL)}

To further enhance the Prediction Agent, we introduce a reinforcement learning phase following SFT. This stage moves beyond pattern imitation to optimize the agent's output for real-world investment objectives. We employ DAPO~\cite{yu2025dapo}, an advanced policy gradient algorithm derived from GRPO~\cite{shao2024deepseekmath}, to align the agent's generation with key metrics of accuracy and rationale quality.

First, We design a reward function, $\operatorname{Reward}(Y, Y^*)$, to holistically evaluate a generated response $Y = [R_{\text{invest}}, R_{\text{rec}}]$ against its ground-truth $Y^* = [R^*_{\text{invest}}, R^*_{\text{rec}}]$. This reward is a weighted combination of the recommendation rating $R_{\text{rec}}$ accuracy and the quality of the investment analysis $R_{\text{invest}}$ measured by ROUGE~\cite{lin2004rouge}. It is defined as follows:
\begin{align*}
\operatorname{Reward}(Y,Y^*) ={}& \alpha \cdot \text{ACC}(R_{\text{rec}}, R^*_{\text{rec}}) \\
            & + \beta \cdot \operatorname{ROUGE-1}(R_{\text{invest}}, R^*_{\text{invest}}) \\
            & + \gamma \cdot \operatorname{ROUGE-L}(R_{\text{invest}}, R^*_{\text{invest}}),
\end{align*}
where $\alpha$, $\beta$, and $\gamma$ are hyperparameters that balance the importance of each component. In our configuration, we set these parameters to $\alpha = 0.6$, $\beta = 0.2$, and $\gamma = 0.2$. This setup is designed to prioritize recommendation accuracy while also considering the quality of the analytical content.

The DAPO algorithm then optimizes the policy $\pi_{\theta}$ by maximizing a clipped surrogate objective, a principle inherited from PPO~\cite{schulman2017proximal} to ensure stable training. The objective can be conceptually expressed as:
$$
    \mathcal{J}_{\text{DAPO}}(\theta) \approx \mathbb{E} \left[ \min \left( r(\theta) \hat{A}, \text{clip}(r(\theta), 1-\epsilon_l, 1+\epsilon_h) \hat{A} \right) \right].
$$
Here, $r(\theta)$ is the probability ratio between the new and old policies, and $\hat{A}$ represents the standardized advantage of a generated sequence. This objective encourages updates that improve rewards while penalizing large policy shifts. The full, detailed formulation is provided in the Appendix.

\subsection{Evaluation System}

To comprehensively evaluate the generated ERRs, we devise a comprehensive evaluation system, which along with the previously constructed FinRpt dataset, forms a complete benchmark.

\paragraph{Basic Metrics}

Basic metrics are used to evaluate the generated ERRs from the perspective of text similarity and prediction accuracy. (1) CompletionRate:
reveals the proportion of cases where the method successfully generates the ERR in the required format.
(2) Accuracy: evaluates the accuracy of the generated recommendation rating (buy or sell). (3) BERTScore~\cite{zhang2019bertscore}: evaluates semantic similarity using BERT embeddings. (4) ROUGE-L~\cite{lin2004rouge}:
one of the ROUGE metric family, which is commonly used for evaluating the quality of summaries, text generation, and machine translation. (5) NumberRate: measures the size of the mathematical number in the generated report $N_{gen}$ to that in the reference ERR $N_{ref}$, revealing the richness of numerical content in the generated report. It is calculated by $\text{NumberRate} = \min(N_{gen}/N_{ref}, 1)$. 

\paragraph{LLM Evaluations}
\label{para:metrics}
To assess the generated ERRs from the aspects of semantic meaning and context, we developed a set of metrics by referring to relevant academic research~\cite{penman2013financial, greenwald2020value} and industry practices: (1) Financial Numeric (FN): evaluates the precision of the data presented and the depth of the financial analysis in the report. (2) News: assesses how relevant and comprehensive the news analysis is in relation to the company and its stock performance. (3) Company \& Market \& Industry (CMI): measures the model's insight into the company's management structure, development trajectory, market trends, and overall industry environment. (4) Invest: evaluates whether the investment recommendations are grounded in thorough, logical, and well-reasoned analysis. (5) Risk: assesses how thoroughly the report analyzes the potential risks associated with investing in the stock. (6) Writing: measures overall coherence, readability, and logical consistency. 

\begin{table*}[htbp] 
 \centering 
 \renewcommand{\arraystretch}{1.1} 
     \resizebox{\textwidth}{!}{
 \begin{tabular}{@{}lcccccc@{}} 
 \toprule 
 \textbf{Category} & \textbf{Method} & \textbf{CompletionRate} & \textbf{Accuracy} & \textbf{ROUGE-L} & \textbf{BERTScore} & \textbf{NumberRate} \\ 
 \midrule 
 \multirow{13}{*}{\textbf{Single LLMs}} & XuanYuan-13B-Chat & 100\% & 33\% & 22.55 & 68.21 & 90.32\% \\ 
 & Gemma2-9B & 100\% & 40\% & 24.97 & 70.83 & 38.92\% \\ 
 & Qwen2.5-7B-Instruct & 100\% & 45\% & 28.08 & 72.06 & 82.52\% \\ 
 & Qwen2.5-14B-Instruct & 100\% & 46\% & 29.74 & 74.24 & 83.40\% \\ 
 & Qwen2.5-72B-Instruct & 100\% & 46\% & 30.17 & 73.77 & 90.00\% \\ 
 & Llama3.1-8B-Instruct & 100\% & 45\% & 26.77 & 71.63 & 68.06\% \\ 
 & Llama3.1-70B & 100\% & 46\% & 29.09 & 72.99 & 76.39\% \\ 
 & GLM4-9B-Chat & 100\% & 45\% & 28.14 & 73.20 & 60.67\% \\ 
 & GPT-4o & 100\% & 48\% & 40.72 & 79.57 & 95.72\% \\ 
 & GPT-4o-mini & 100\% & 47\% & 39.45 & 78.89 & 95.75\% \\ 
 & Gemini-2.5-Pro & 100\% & 50\% & 41.79 & 80.29 & 87.23\% \\ 
 \midrule 
 \multirow{8}{*}{\textbf{\makecell{FinRpt-Gen with\\ open-source LLMs}}} 
 & FinRpt-Gen (Gemma2-9B) & 95\% & 47\% & 32.83 & 72.63 & 80.99\% \\ 
 & FinRpt-Gen (Qwen2.5-7B-Instruct) & 98\% & 48\% & 34.51 & \textit{72.65} & 84.29\% \\ 
 & FinRpt-Gen (Qwen2.5-14B-Instruct) & 100\% & 48\% & 36.27 & 77.05 & 93.16\% \\ 
 & FinRpt-Gen (Qwen2.5-72B-Instruct) & 100\% & 49\% & 36.88 & 76.14 & 94.00\% \\ 
 & FinRpt-Gen (Llama3.1-8B-Instruct) & 93\% & 45\% & 30.03 & 68.38 & 58.92\% \\ 
 & FinRpt-Gen (Llama3.1-70B-Instruct) & 97\% & 48\% & 34.28 & 73.43 & 83.39\% \\ 
 & FinRpt-Gen (GLM4-9B-Chat) & 100\% & 50\% & 38.35 & 76.66 & 76.09\% \\ 
 \midrule 
 \multirow{3}{*}{\textbf{\makecell{FinRpt-Gen with \\ closed-source LLMs}}} & FinRpt-Gen (GPT-4o) & 100\% & 51\% & 48.44 & 82.09 & \textbf{98.62\%} \\ 
 & FinRpt-Gen (GPT-4o-mini) & 100\% & 50\% & 44.09 & 80.82 & 97.01\% \\ 
 & FinRpt-Gen (Gemini-2.5-Pro) & 100\% & 51\% & 48.58 & 82.12 & 90.57\% \\ 
 \midrule 
 \multirow{3}{*}{\textbf{\makecell{FinRpt-Gen with \\fine-tuned LLMs}}} & FinRpt-Gen (Llama3.1-8B-Instruct-SFT) & 100\% & 50\% & 48.67 & 82.14 & 94.14\% \\ 
 & FinRpt-Gen (GLM4-9B-Chat-SFT) & 100\% & 51\% & 48.64 & 82.16 & 93.86\% \\ 
 & FinRpt-Gen (Qwen2.5-7B-Instruct-SFT) & 100\% & 54\% & 48.83 & 82.21 & 94.48\% \\ 
 \midrule 
 \multirow{1}{*}{\textbf{Our}} & FinRpt-Gen (Qwen2.5-7B-Instruct-SFT-RL) & \textbf{100\%} & \textbf{55\%} & \textbf{49.06} & \textbf{82.43} & 95.15\% \\ 
 \bottomrule 
 \end{tabular} 
 }
\caption{Performance comparison of FinRpt-Gen against baselines under the evaluation of basic metrics.}
  \label{tab:basic_results}%
\end{table*} 

We leverage a conservative
approach to compare the performance of LLMs following previous works~\cite{zheng2023judging, liu2024aligning, fu2023gptscore}. For each sample, the ERRs generated by different models are compared pairwise using a Judge Agent (GPT-4o). To eliminate position bias, the judge evaluates the ERRs twice, with their order swapped. A win is recorded only if one answer is preferred in both orders; otherwise, the result is marked as a tie. Once the judging process is finished, we could calculate the Win Counts, the Tie Counts, and the Loss Counts. We further calculate the Adjusted Win Rate for the sake of comparison:

{\small
\begin{equation*}
    \begin{aligned}
        \text{Adjusted Win Rate}& = \\
        & \frac{\text{Win Counts} + 0.5 \cdot \text{Tie Counts}}{\text{Win Counts} + \text{Loss Counts} + \text{Tie Counts}}.
    \end{aligned}
\end{equation*}
}

\section{Experiments}

\subsection{Experiment Setting}
\subsubsection{Implement Detail}
The open-source models were accessed via the Ollama Python Library locally, while the closed-source models were accessed through their official APIs. The SFT phase was conducted on 8 NVIDIA 3090 GPUs, and the RL phase used 8 NVIDIA A100 GPUs. We randomly selected 100 samples from the FinRpt test set for evaluation. For detailed hyperparameters and further implementation specifics, please refer to the Appendix.

\subsubsection{Baselines}

We evaluate our method against four types of baselines: (1) standalone state-of-the-art LLMs, (2) our FinRpt-Gen framework with closed-source LLMs, (3) the framework with open-source LLMs, and (4) the framework with fine-tuned open-source LLMs. Please see the Appendix for a more detailed description of all baselines.

\subsection{Main Results}

\subsubsection{Main Results of Basic Metrics}

We compare the performance of FinRpt-Gen against strong baselines, and the results are shown in Table~\ref{tab:basic_results}, from which we can draw the following conclusions: The performance of the multi-agent framework FinRpt-Gen is significantly better than that of single LLMs, which highlights the effectiveness of our multi-agent framework. 
In the case without SFT and RL,
the performance of the closed-source models Gemini-2.5-Pro and GPT-4o is better than the selected open-source models with a clear margin. This is an expected outcome, as Gemini-2.5-Pro and GPT-4o are widely recognized as leading models in the field. After applying SFT on the constructed dataset FinRpt, there is an obvious improvement in performance compared to the results without SFT,
and it even outperforms these two closed-source models in almost all evaluation aspects. After further enhancement through RL, optimal performance is achieved. This partially reflects the high quality of our dataset.

\begin{figure}[h]
 \centering
\includegraphics[width=\linewidth]{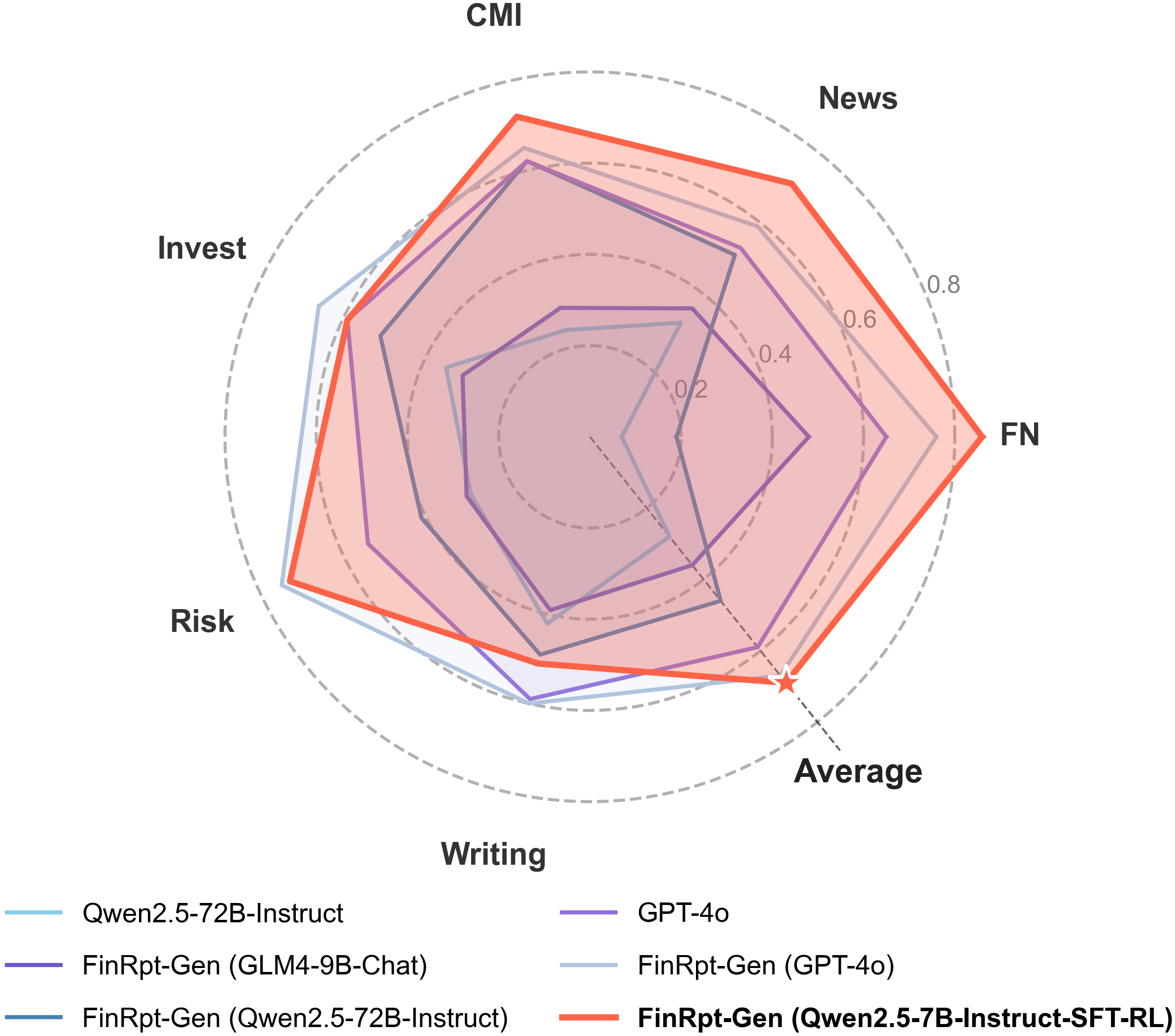}
\caption{The performance comparison under the LLM evaluation metrics.}
\label{fig:llm_eval_radar}
\end{figure}

\subsubsection{Main Results of LLM Evaluations}

Based on the LLM evaluation metrics previously detailed, we compare the performance of models from a financial professionalism perspective. The results are shown in Figure~\ref{fig:llm_eval_radar}. And the detailed quantitative results are available in Table~\ref{tab:llm_evaluation_metrics} in the Appendix.
This radar chart illustrates our trained model achieves excellent performance comparable to GPT-4o and surpasses all other strong baselines. Notably, our trained model even excels FinRpt-Gen (GPT-4o) in CMI, News and FN metrics. By comparing the results,
 we can also conclude the effectiveness of the FinRpt-Gen framework and the training dataset FinRpt. To further validate the reliability of our LLM evaluation, we conducted a human evaluation study in Appendix.

\subsubsection{Resource Requirements Analysis}

The framework's resource requirements are minimal. The entire process of generating an ERR, from data crawling to report creation, is completed in approximately 3 to 4 minutes. For a detailed breakdown of the resource requirements, including processing times and API costs, please refer to the Appendix.

\subsection{Ablation Study}
To demonstrate the effectiveness of the components used in FinRpt-Gen (Qwen2.5-7B-Instruct-SFT-RL), abbreviated as FinRpt-Gen in this section, we compare it with four variants: (1) FinRpt-Gen w/o Finance Extraction: removing Income, Balance, and Cash Extraction Agents and inputting the corresponding information into the Information Analysis Module directly. (2) FinRpt-Gen w/o News Extraction: removing the News Extraction Agent and inputting news data into the Information Analysis Agent directly. (3) FinRpt-Gen w/o 3 Analysis Agents: replacing the Finance, Status, and Risks Analysis Agent with one single LLM (GPT-4o).

\begin{table}[htbp]
  \centering
  \resizebox{\linewidth}{!}{
    \begin{tabular}{cccc}
    \toprule
    \textbf{Method} & \textbf{Accuracy} & \textbf{ROUGE-L} & \textbf{BERTScore} \\
    \midrule
    w/o Finance Extraction & 47    & 38,93 & 76.50 \\
    w/o News Extraction & 49    & 46.02 & 81.20 \\
    w/o 3 Analysis Agents & 51    & 45.92 & 81.38 \\
    \textbf{FinRpt-Gen} & \textbf{55}    & \textbf{49.06} & \textbf{82.43} \\
    \bottomrule
    \end{tabular}}%
      \caption{Ablation study results over 3 variants.}
  \label{tab:ablation}%
\end{table}%

The ablation results are shown in Table~\ref{tab:ablation}, from which we can see that FinRpt-Gen outperforms other variants with clear advantages. The performance of FinRpt-Gen drops significantly without the Finance Extraction Agent, 
highlighting the necessity of this agent. Similarly, the absence of the News Extraction Agent leads to a noticeable decline in performance, demonstrating the value of pre-extracting and summarizing key news of the agent. Furthermore, utilizing three Analysis Agents outperforms relying on a single GPT-4 model, confirming the importance of the specialized design of each Analysis Agent.

\subsection{Dataset Quality Study}

\begin{table}[htbp]
  \centering
  \resizebox{\linewidth}{!}{
    \begin{tabular}{ccccc|c}
    \toprule
        \textbf{ERRs} & \textbf{FN} & \textbf{News} & \textbf{Invest} & \multicolumn{1}{c}{\textbf{Writing}} & \textbf{Average} \\
    \midrule
    Expert-written & 4.57  & 4.00  & 4.13  & 4.33  & 4.30 \\
    FinRpt & 4.33  & 4.20  & 4.10 & 4.17  & 4.20 \\
    Kappa Score & 0.85  & 0.86 & 0.89 & 0.84  & 0.86 \\
    \bottomrule
    \end{tabular}%
    }
      \caption{Comparing the quality of ERRs in dataset FinRpt and expert-written ERRs.}
  \label{tab:human_evaluation}%
\end{table}%

\subsubsection{Human Evaluation}

In the Data Construction Pipeline, the Dataset Enhancement Module is devised to enhance the data quality. To investigate the dataset quality thoroughly, we also conduct a human evaluation. We randomly sample 30 ERRs from the FinRpt dataset and 30 ERRs written by experts. Then three senior financial analysts are invited to rate each ERR from four aspects as devised in LLM Evaluations Section with a scoring range of 0 to 5. These three senior analysts are carefully chosen based on their extensive experience and expertise in the field, ensuring they can provide reliable assessments. We also provide them with detailed evaluation guidelines and criteria to ensure consistency in their judgments. As the results presented in Table~\ref{tab:human_evaluation}, the scores of FinRpt and expert-written are very close, suggesting a high level of data quality. Besides, we categorize the evaluation scores of each ERR into three classes and calculate the Fleiss’ kappa score~\cite{landis1977measurement} (ranging from -1 to 1) to evaluate agreement among the evaluators. The kappa scores are shown in the last line of Table~\ref{tab:human_evaluation}, confirming the consistency and reliability of the human evaluation.

\subsubsection{Case Study}

We randomly select an ERR case from FinRpt dataset.
The case is presented in
Figure \ref{fig:en_report} in Appendix.
This case shows the high quality of our dataset from the following four perspectives. (1) The detailed quantitative financial metrics underscore the report's accuracy and thoroughness in financial analysis. (2) The forward-looking strategic analysis offers investors valuable insights into long-term growth drivers and potential risks. (3) The clear investment thesis, supported by quantitative data and strategic context, reflects a balanced approach to stock valuation. (4) The well-structured format ensures clarity and easy navigation, enhancing the report's usability for investors.
Collectively, these perspectives highlight the high quality of our dataset, underscoring its utility as a valuable resource for the ERR generation task and other Fintech fields.

\section{Conclusion}

This paper formulated the Equity Research Report generation task, and proposed an open-source benchmark FinRpt consisting of a high-quality ERR dataset and a comprehensive evaluation system designed to assess various aspects of ERR generation. Additionally, we tailored a multi-agent framework FinRpt-Gen for this task by applying SFT and RL to our proposed datasets. Experimental results demonstrate the data quality, metric effectiveness of benchmark FinRpt, and strong performance of FinRpt-Gen, highlighting its potential to advance ERR generation. 

\bibliography{custom}

\clearpage

\appendix

\section{Data Collection Module Supplement}

The detailed content of each data type and data source of the Data Collection Module is present in Table~\ref{tab:data}.

\begin{table*}[htbp]
  \centering
    \begin{threeparttable}
      \resizebox{0.73\textwidth}{!}{
      \begin{tabular}{lll}
      \toprule
      \textbf{Data Type} & \textbf{Introduction} & \textbf{Data Source} \\
      \midrule
      Company Information $O$ & \makecell[l]{Encompassing its country of operation,\\ management structure, market positioning, \\and industry classification.} & \makecell[l]{Yahoo Finance (API)\\ finance.yahoo.com} \\
      \midrule
      Financial Indicators $F$ & \makecell[l]{Including the income statements, balance \\ sheets, and cash flow statements.} & \makecell[l]{AKShare (API)\\ akshare.akfamily.xyz} \\
      \midrule
      Company Announcements $A$ & \makecell[l]{Including board resolution announcements,\\supervisory board resolution announcements, \\annual, quarterly, and audit reports.} & \makecell[l]{Eastmoney (Crawling)\\ eastmoney.com} \\
      \midrule
      Company-related News $N$ & \makecell[l]{Including the company-related news.} & \makecell[l]{Sina Finance (Crawling)\\ finance.sina.com.cn} \\
      \midrule
      Historical Stock Prices $P$ & \makecell[l]{The historical close price.}   & AKShare (API)\\
      \midrule
      Historical Market Indices $M$ & \makecell[l]{The CSI300 Index of the Chinese market.} & AKShare (API) \\
      \bottomrule
      \end{tabular}}
    \end{threeparttable}%
      \caption{The introduction of data contents and data sources corresponding to 6 data types.}
  \label{tab:data}%
\end{table*}%

\section{LLM Evaliation Results}
\label{app:LLM_evaluation}

We compare the performance of models from a financial professionalism perspective and the results are shown in Table~\ref{tab:llm_evaluation_metrics}.

\begin{table*}[h]
  \centering
  \resizebox{\textwidth}{!}{
    \begin{tabular}{cccccccc|c}
    \toprule
    \textbf{Method} & \textbf{Criteria} & \textbf{FN} & \textbf{News} & \textbf{CMI} & \textbf{Invest} & \textbf{Risk} & \multicolumn{1}{c}{\textbf{Writing}} & \textbf{Average} \\
    \midrule
    \multirow{6}[2]{*}{Qwen2.5-72B-Instruct } & Win   & 5     & 12    & 20    & 24    & 30    & 25    & 19.33 \\
          & Loss  & 263   & 117   & 173   & 112   & 156   & 73    & 149.00 \\
          & Tie   & 30    & 170   & 106   & 162   & 113   & 201   & 130.33 \\
          & Win Rate & 0.02  & 0.04  & 0.07  & 0.08  & 0.10  & 0.08  & 0.07 \\
          & Loss Rate & 0.88  & 0.39  & 0.58  & 0.38  & 0.52  & 0.24  & 0.50 \\
          & Adjusted Win Rate  & 0.07  & 0.32  & 0.24  & 0.35  & 0.29  & 0.42  & 0.28 \\
    \midrule
    \multirow{6}[2]{*}{FinRpt-Gen (GLM4-9B-Chat)} & Win   & 114   & 52    & 40    & 36    & 48    & 40    & 55.00 \\
          & Loss  & 124   & 136   & 166   & 151   & 168   & 107   & 142.00 \\
          & Tie   & 57    & 108   & 90    & 108   & 80    & 149   & 98.67 \\
          & Win Rate & 0.39  & 0.18  & 0.14  & 0.12  & 0.16  & 0.14  & 0.19 \\
          & Loss Rate & 0.42  & 0.46  & 0.56  & 0.51  & 0.57  & 0.36  & 0.48 \\
          & Adjusted Win Rate & 0.48  & 0.36  & 0.29  & 0.31  & 0.30  & 0.39  & 0.36 \\
    \midrule
    \multirow{6}[2]{*}{FinRpt-Gen (Qwen2.5-72B-Instruct)} & Win   & 32    & 57    & 86    & 63    & 63    & 58    & 59.83 \\
          & Loss  & 206   & 52    & 92    & 57    & 114   & 63    & 97.33 \\
          & Tie   & 40    & 168   & 109   & 159   & 102   & 166   & 124.00 \\
          & Win Rate & 0.12  & 0.21  & 0.30  & 0.23  & 0.23  & 0.20  & 0.22 \\
          & Loss Rate & 0.74  & 0.19  & 0.14  & 0.20  & 0.41  & 0.22  & 0.32 \\
          & Adjusted Win Rate  & 0.19  & 0.51  & 0.62  & 0.51  & 0.41  & 0.49  & 0.46 \\
    \midrule
    \multirow{6}[2]{*}{GPT-4o} & Win   & 150   & 83    & 109   & 102   & 107   & 85    & 106.00 \\
          & Loss  & 63    & 68    & 40    & 51    & 85    & 31    & 56.33 \\
          & Tie   & 80    & 136   & 130   & 134   & 101   & 170   & 125.17 \\
          & Win Rate & 0.51  & 0.29  & 0.39  & 0.36  & 0.37  & 0.30  & 0.37 \\
          & Loss Rate & 0.22  & 0.24  & 0.14  & 0.18  & 0.29  & 0.11  & 0.20 \\
          & Adjusted Win Rate  & 0.65  & 0.53  & 0.62  & 0.59  & 0.54  & 0.59  & 0.59 \\
    \midrule
    \multirow{6}[2]{*}{FinRpt-Gen (GPT-4o)} & Win   & 182   & 101   & 141   & 129   & 174   & 96    & 137.17 \\
          & Loss  & 34    & 50    & 50    & 36    & 31    & 35    & 39.33 \\
          & Tie   & 67    & 141   & 101   & 121   & 81    & 162   & 112.17 \\
          & Win Rate & 0.64  & 0.35  & 0.48  & 0.45  & 0.61  & 0.33  & 0.48 \\
          & Loss Rate & 0.12  & 0.17  & 0.17  & 0.13  & 0.11  & 0.12  & 0.14 \\
          &  Adjusted Win Rate & 0.76  & 0.59  & 0.65  & \textbf{0.66} & \textbf{0.75} & \textbf{0.60} & 0.67 \\
    \midrule
    \multirow{6}[2]{*}{FinRpt-Gen (Qwen2.5-7B-Instruct-SFT-RL) } & Win   & 218   & 147   & 158   & 110   & 166   & 40    & 139.83 \\
          & Loss  & 11    & 29    & 33    & 57    & 34    & 35    & 33.17 \\
          & Tie   & 56    & 109   & 94    & 126   & 87    & 204   & 112.67 \\
          & Win Rate & 0.76  & 0.52  & 0.55  & 0.38  & 0.58  & 0.14  & 0.49 \\
          & Loss Rate & 0.04  & 0.10  & 0.12  & 0.19  & 0.12  & 0.13  & 0.12 \\
          & Adjusted Win Rate & \textbf{0.86} & \textbf{0.71} & \textbf{0.72} & 0.59  & 0.73  & 0.51  & \textbf{0.69} \\
    \bottomrule
    \end{tabular}%
    }
      \caption{The performance comparison under the LLM evaluation metrics.}
  \label{tab:llm_evaluation_metrics}%
\end{table*}%

\section{Detailed Implement Setting}
LLM Access and Inference: For all LLMs, the temperature was set to 0, top p to 1, and both frequency penalty and presence penalty were set to 0.
Supervised Fine-Tuning: The SFT phase was conducted over 3 epochs with a batch size of 1 per device. We employed a cosine learning rate scheduler with a learning rate of 1.0e-4. LoRA was applied to all linear modules with a rank of 8 and an alpha of 16.
Reinforcement Learning: We utilized the DAPO algorithm for 5 epochs with a learning rate of 1.0e-6 and a batch size of 64. In the policy rollout stage, the temperature was set to 1.0 and top p to 1.0, generating 16 responses for each prompt. The low and high clip ratios were set to 0.2 and 0.28, respectively.

\section{Detailed Baselines}
Our evaluation incorporates four distinct categories of baselines. The first category consists of standalone state-of-the-art LLMs, including XuanYuan-13B-Chat~\cite{zhang2023xuanyuan}, Gemma2-9B~\cite{team2024gemma}, various configurations of Qwen2.5-Instruct (7B, 14B, and 72B)~\cite{qwen2.5}, Llama3.1-Instruct (8B, 70B)~\cite{dubey2024llama}, GLM4-9B-Chat~\cite{glm2024chatglm}, GPT-4o~\cite{hurst2024gpt} and its mini version, and Gemini-2.5-Pro~\cite{comanici2025gemini}. The second category features our FinRpt-Gen framework integrated with closed-source LLMs, specifically with GPT-4o, GPT-4o-mini, and Gemini-2.5-Pro. The third group comprises the FinRpt-Gen framework with open-source LLMs, such as Gemma2-9B, different Qwen2.5-Instruct models (7B, 14B, 72B), Llama3.1-8B-Instruct, Llama3.1-70B-Instruct, and GLM4-9B-Chat. Lastly, the fourth category evaluates the FinRpt-Gen framework when combined with fine-tuned open-source LLMs, namely Llama3.1-8B-Instruct-SFT, GLM4-9B-Chat-SFT, and Qwen2.5-7B-Instruct-SFT.

\section{Detailed Resource Requirements Analysis}
\begin{figure}[h]
 \centering
\includegraphics[width=\linewidth]{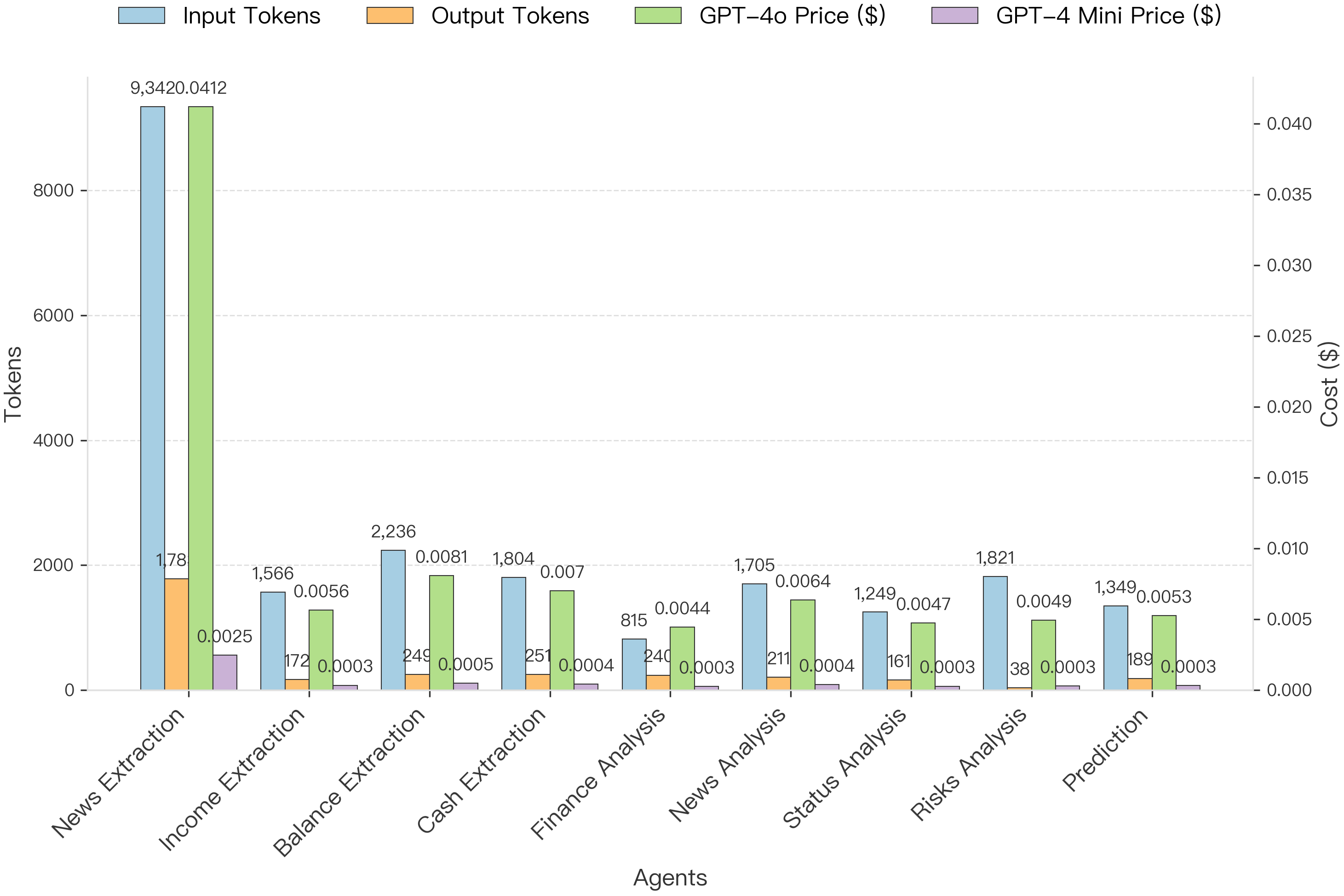}
\caption{ The
average input tokens, output tokens, and costs for each agent. (The price calculation is based on the official API price of OpenAI as of February 1, 2025.)}
\label{fig:resource_requirements_analysis}
\end{figure}

The framework's resource consumption is broken down into the following steps.
Data Crawling: The initial step of gathering relevant company information requires 120-160 seconds. This process can be executed in advance, and the collected data can be cached for faster retrieval.
ERR Generation: It takes approximately 60 seconds to make 9 API calls to generate the final ERR.
Report Creation: The framework utilizes the ReportLab Python Library to create the PDF version of the ERR, a process that takes about 1 second.
The average input tokens, output tokens, and associated costs for each agent are summarized in Figure~\ref{fig:resource_requirements_analysis}. The cost calculations are based on the official OpenAI API pricing as of February 1, 2025.

\section{Report Case}
\label{app:report_case}

\begin{figure*}[h]
 \centering
\includegraphics[width=0.84\linewidth]{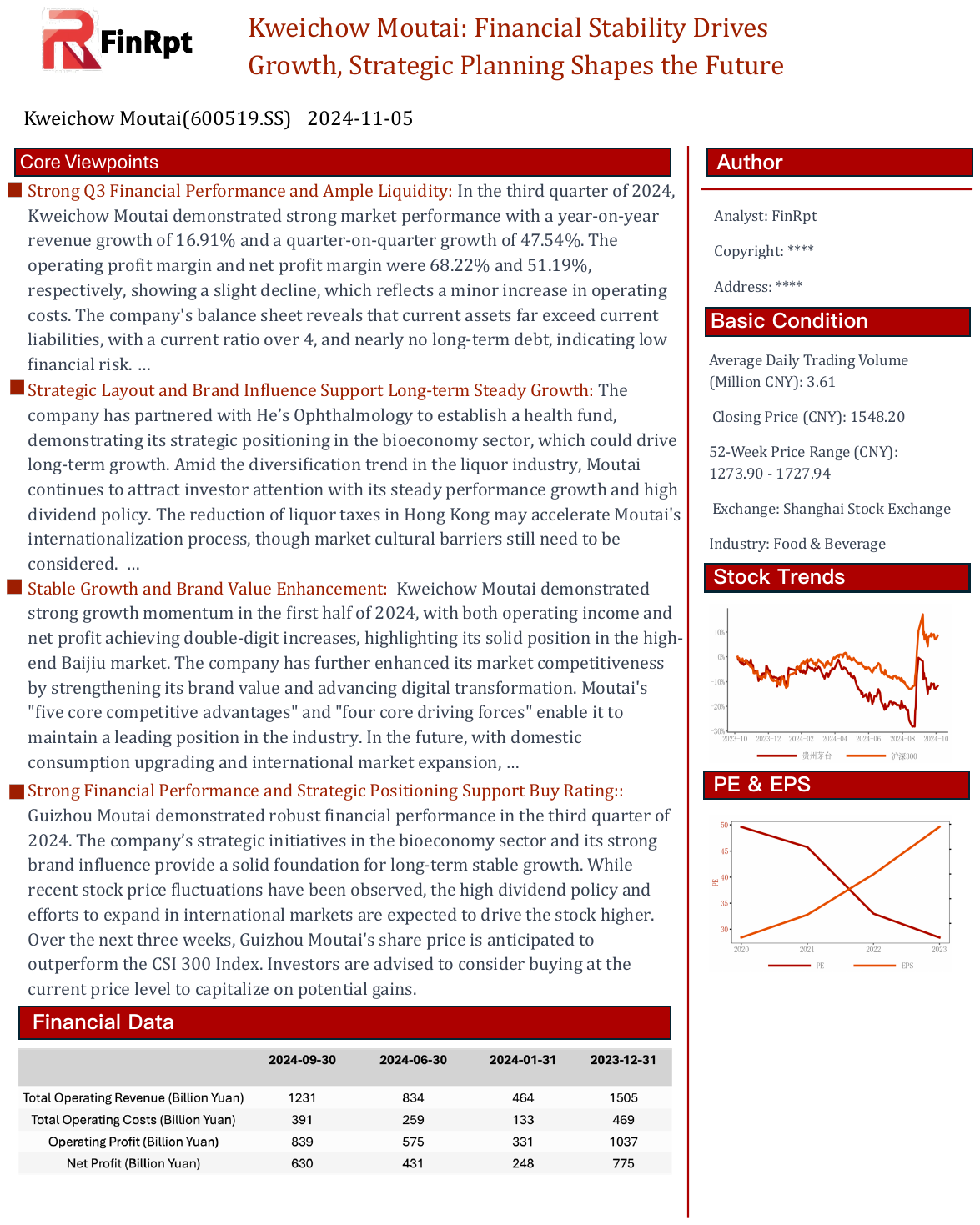}
\caption{An ERR case with the stock ticker 600519.SS at analysis date 2024-11-05.}
\label{fig:en_report}
\end{figure*}

We randomly selected a ERR case from FinRpt dataset, shown in Figure \ref{fig:en_report}.

\section{Detailed DAPO Formulation}

We provide the full mathematical formulation of the Dynamic sAmpling Policy Optimization (DAPO) objective function here. The goal is to maximize the expected clipped reward over a batch of generated sequences.

The DAPO objective, $\mathcal{J}_{\text{DAPO}}(\theta)$, is defined as:
\begin{align*}
    \mathcal{J}_{\text{DAPO}}(\theta) &= \mathbb{E}_{(q,a) \sim \mathcal{D}, \{o_i\}_{i=1}^G \sim \pi_{\theta_{\text{old}}}(\cdot|q)} \Bigg[ \\
    & \quad \frac{1}{\sum_{i=1}^G |o_i|} \sum_{i=1}^G \sum_{t=1}^{|o_i|} \min \Big( r_{i,t}(\theta) \hat{A}_{i,t}, \\
    & \quad \qquad \text{clip}(r_{i,t}(\theta), 1-\epsilon_{\text{low}}, 1+\epsilon_{\text{high}}) \hat{A}_{i,t} \Big)
    \Bigg],
\end{align*}
where $(q,a) \sim \mathcal{D}$ represents a prompt-answer pair sampled from the dataset, $\{o_i\}_{i=1}^G$ is a set of $G$ output sequences generated from the prompt $q$ using the old policy $\pi_{\theta_{\text{old}}}$, and  $|o_i|$ is the length of the $i$-th sequence. The objective is normalized by the total number of tokens across all $G$ generations in the batch.

The key components of the objective are the per-token probability ratio $r_{i,t}(\theta)$ and the standardized advantage estimate $\hat{A}_{i,t}$.

The probability ratio $r_{i,t}(\theta)$ compares the likelihood of generating the $t$-th token of sequence $o_i$ under the new policy $\pi_{\theta}$ versus the old policy $\pi_{\theta_{\text{old}}}$:
\[
r_{i,t}(\theta) = \frac{\pi_{\theta}(o_{i,t} \mid q, o_{i,<t})}{\pi_{\theta_{\text{old}}}(o_{i,t} \mid q, o_{i,<t})}.
\]

The standardized advantage $\hat{A}_{i,t}$ is calculated at the sequence level and is constant for all tokens $t$ within the same sequence $o_i$. It measures how much better the reward for sequence $o_i$ is compared to the average reward of other sequences in the batch:
\[
\hat{A}_{i,t} = \frac{R_i - \text{mean}(\{R_j\}_{j=1}^G)}{\text{std}(\{R_j\}_{j=1}^G)},
\]
where, $R_i$ is the reward for the complete sequence $o_i$, calculated using the reward function defined in the main text. The mean and standard deviation are computed over the rewards of all $G$ sequences in the batch.

\section{Human Evaluation Study for LLM Evaluation}
To validate the reliability of our LLM-based evaluation, we conducted a human verification study. A subset of 50 distinct pairs was randomly sampled from the full set evaluated by our GPT-4o Judge Agent. These pairs were then assessed by three human evaluators, all of whom are senior financial analysts. The evaluators were instructed to independently determine which sample in each pair was superior, following the exact same guidelines and criteria provided to the Judge Agent. A majority vote was used to establish the definitive human-preferred sample for each pair.
We then compared this human consensus against the decisions made by the GPT-4o Judge Agent. As shown in Table~\ref{tab:human_study}, the results indicate a high degree of concordance. The LLM's judgments aligned with the human majority vote in 45 out of 50 cases, achieving a 90\% agreement rate. This strong correlation between our automated and human assessments lends significant credibility to our LLM-based evaluation methodology.

\begin{table*}[htbp]
  \centering
    \resizebox{\textwidth}{!}{
    \begin{tabular}{ccccccc}
    \toprule
    \textbf{Pair ID} & \textbf{Evaluator 1} & \textbf{Evaluator 2} & \textbf{Evaluator 3} & \textbf{Human Consensus (Majority)} & \textbf{GPT-4o Judge Agent} & \textbf{Agreement (Human vs. GPT-4o)} \\
    \midrule
    \textbf{1} & A     & A     & A     & A     & A     & Yes \\
    \textbf{2} & B     & A     & B     & B     & B     & Yes \\
    \textbf{3} & A     & B     & A     & A     & A     & Yes \\
    \textbf{4} & B     & B     & B     & B     & B     & Yes \\
    \textbf{5} & A     & A     & B     & A     & A     & Yes \\
    \textbf{6} & B     & B     & A     & B     & A     & No \\
    \textbf{7} & A     & A     & A     & A     & A     & Yes \\
    \textbf{8} & A     & B     & B     & B     & B     & Yes \\
    \textbf{9} & A     & B     & A     & A     & A     & Yes \\
    \textbf{10} & B     & A     & B     & B     & B     & Yes \\
    \textbf{11} & A     & A     & B     & A     & A     & Yes \\
    \textbf{12} & B     & B     & A     & B     & B     & Yes \\
    \textbf{13} & A     & A     & A     & A     & A     & Yes \\
    \textbf{14} & B     & B     & B     & B     & B     & Yes \\
    \textbf{15} & B     & A     & B     & B     & A     & No \\
    \textbf{16} & B     & A     & B     & B     & B     & Yes \\
    \textbf{17} & A     & A     & B     & A     & A     & Yes \\
    \textbf{18} & B     & B     & A     & B     & B     & Yes \\
    \textbf{19} & A     & A     & A     & A     & A     & Yes \\
    \textbf{20} & B     & B     & B     & B     & B     & Yes \\
    \textbf{21} & A     & B     & A     & A     & A     & Yes \\
    \textbf{22} & B     & A     & B     & B     & B     & Yes \\
    \textbf{23} & A     & A     & B     & A     & B     & No \\
    \textbf{24} & B     & B     & A     & B     & B     & Yes \\
    \textbf{25} & A     & A     & A     & A     & A     & Yes \\
    \textbf{26} & B     & A     & B     & B     & B     & Yes \\
    \textbf{27} & A     & A     & B     & A     & A     & Yes \\
    \textbf{28} & B     & B     & B     & B     & B     & Yes \\
    \textbf{29} & A     & B     & A     & A     & B     & No \\
    \textbf{30} & A     & B     & A     & A     & A     & Yes \\
    \textbf{31} & A     & A     & B     & A     & A     & Yes \\
    \textbf{32} & B     & B     & A     & B     & B     & Yes \\
    \textbf{33} & A     & A     & A     & A     & A     & Yes \\
    \textbf{34} & B     & B     & B     & B     & B     & Yes \\
    \textbf{35} & A     & B     & A     & A     & A     & Yes \\
    \textbf{36} & B     & A     & B     & B     & B     & Yes \\
    \textbf{37} & A     & A     & B     & A     & A     & Yes \\
    \textbf{38} & B     & B     & A     & B     & B     & Yes \\
    \textbf{39} & A     & A     & A     & A     & B     & No \\
    \textbf{40} & B     & B     & B     & B     & B     & Yes \\
    \textbf{41} & A     & B     & A     & A     & A     & Yes \\
    \textbf{42} & B     & A     & B     & B     & B     & Yes \\
    \textbf{43} & A     & A     & B     & A     & A     & Yes \\
    \textbf{44} & B     & B     & A     & B     & B     & Yes \\
    \textbf{45} & A     & A     & A     & A     & A     & Yes \\
    \textbf{46} & B     & B     & B     & B     & B     & Yes \\
    \textbf{47} & A     & B     & A     & A     & A     & Yes \\
    \textbf{48} & B     & A     & B     & B     & B     & Yes \\
    \textbf{49} & A     & A     & B     & A     & A     & Yes \\
    \textbf{50} & B     & B     & A     & B     & B     & Yes \\
    \midrule
    \textbf{Total} &       &       &       &       & \textbf{Agreements:} & \textbf{45 / 50 (90.0\%)} \\
    \bottomrule
    \end{tabular}%
    }
    \caption{Detailed human evaluation study results.}
  \label{tab:human_study}%
\end{table*}%

\section{Prompts}
\label{app:prompt}

For each agent, we present its prompt and an input example in the following tables: 
News Extraction Agent in Table \ref{tab:news_extraction_prompt}, Income Extraction Agent in Table \ref{tab:income_extraction_prompt}, Balance Extraction Agent in Table \ref{tab:balance_extraction_prompt}, Cash Extraction Agent in Table \ref{tab:cash_extraction_prompt}, Finance Analysis Agent in Table \ref{tab:finance_ana_prompt}, News Analysis Agent in Table \ref{tab:news_ana_prompt},  Status Analysis Agent in Table \ref{tab:status_ana_prompt}, Risks Analysis Agent in Table \ref{tab:status_ana_prompt} and Prediction Agent in Table \ref{tab:prediction_prompt}.

\begin{table*}[h]
    \centering

    \begin{tabular}{p{15cm}}
    \toprule
    \textbf{[Analysis Date]}:
    2024-11-05 \\
    \\
    \textbf{[Company Name]}:
    Kweichow Moutai Co.,Ltd.\\
    \\
    \textbf{[Recent News]}:\\
    News 1: \\
    News Date: 2024-11-03\\
    News Title: Where is the breakthrough point for Zhang Xu's new position?\\
    News Summary: On November 1, Zhang Xu assumed the post of Secretary of the Party Committee and Chairman of the Board of Directors of Kweichow Moutai Liquor Sales Company, marking a new stage of his career in Maotai. Zhang Xu emphasized that he will continue to pay attention to frontline market voices, stabilize prices, enhance market confidence, and be committed to promoting the healthy and sustainable development of Maotai. His appointment also means that Maotai will respond to marketing challenges with reform and innovative thinking.\\
    News 2: \\
    News Date: 2024-10-30\\
    News Title: Moutai Group deputy general manager Tu Huabin visited Central Control Technology. \\
    News Summary: On October 29, the senior management team of Moutai Group visited Central Control Technology, and the two sides had in-depth exchanges on the intelligent upgrading of the Baijiu industry. Tu Huabin stated that Maotai will integrate traditional brewing techniques with new technologies. Cui Shan, chairman of Central Control Technology, emphasized that the company looks forward to deepening cooperation with Maotai Group. The two sides discussed the intelligent factory structure and specific implementation path to lay the foundation for future cooperation.\\
    ...\\
    News 60:\\
    News Date: 2024-10-14 \\
    News Title: Maotai Group held the Maotai Liquor Festival in Guizhou. \\
    News summary: On October 11, 2024, the Moutai Group held the ``Jiachen'' Year Moutai Wine Festival in Guizhou. During the event, Moutai practitioners carried out an apprenticeship ceremony to pass on winemaking skills, demonstrating their respect and appreciation for traditional culture. Additionally, the festival featured an intangible cultural heritage market, showcasing Guizhou's traditional crafts and cuisine, promoting cultural exchange, and highlighting Moutai wine's value as an intangible cultural heritage. \\
    \\
    \textbf{[Instruction]}: \\
    You are an expert financial analyst specializing in identifying critical news that impacts stock market trends. I provide you with the names of companies and recent news articles above. Your task is to rank the news based on its potential impact on the companies' stock, selecting up to ten of the most important news. Focus on factors such as financial performance, market conditions, regulatory changes, leadership changes, and other relevant information. Limit your output to a maximum of ten key news items, with a total word count not exceeding 3,000 characters. For better readability, format your output in JSON as follows: \texttt{\{``news'': [\{``date'': ``News Date'', ``content'': ``News Content'', ``potential\_impact'': ``Positive/Negative/Neutral''\},\{
      ``date'': ``News Date'',
      ``content'': ``News Content'',
      ``potential\_impact'': ``Positive/Negative/Neutral''\},\ldots,\{
      ``date'': ``News Date'',
      ``content'': ``News Content'',
      ``potential\_impact'': ``Positive/Negative/Neutral''\}]\}} \\
    \bottomrule
    \end{tabular}
    \caption{News Extraction Agent prompt example.}
    \label{tab:news_extraction_prompt}
\end{table*}

\begin{table*}[h]
    \centering

    \begin{tabular}{p{15cm}}
    \toprule
    \textbf{[Analysis Date]}:
    2024-11-05 \\
    \\
    \textbf{[Company Name]}:
    Kweichow Moutai Co., Ltd.\\
    \\
    \textbf{[Income Statement]}: \\
    Date: 2024-09-30, Total Operating Revenue: 123,122,542,625.45, Total Operating Cost: 39,182,396,617.66, Operating Profit: 83,996,733,560.98, Net Profit: 63,031,462,239.55, Basic Earnings per Share: 48.42, Diluted Earnings per Share: 48.42, Investment Income: 8,019,358.73, Quarter-on-Quarter Revenue Growth Rate: 47.54\%, Year-on-Year Revenue Growth Rate: 16.91\%, Gross Profit Margin: 68.18\%, Operating Profit Margin: 68.22\%, Net Profit Margin: 51.19\%; \\
    Date: 2024-06-30, Total Operating Revenue: 83,451,164,646.53, Total Operating Cost: 25,936,116,155.36, Operating Profit: 57,550,951,619.01, Net Profit: 43,176,914,345.12, Basic Earnings per Share: 33.19, Diluted Earnings per Share: 33.19, Investment Income: 2,288,120.31, Quarter-on-Quarter Revenue Growth Rate: 79.52\%, Year-on-Year Revenue Growth Rate: 17.56\%, Gross Profit Margin: 68.92\%, Operating Profit Margin: 68.96\%, Net Profit Margin: 51.74\%; \\
    ... \\
    Date: 2022-12-31, Total Operating Revenue: 127,553,959,355.97, Total Operating Cost: 39,748,309,616.85, Operating Profit: 87,879,521,782.39, Net Profit: 65,376,039,957.88, Basic Earnings per Share: 49.93, Diluted Earnings per Share: 49.93, Investment Income: 63,840,000.00, Quarter-on-Quarter Revenue Growth Rate: 42.06\%, Year-on-Year Revenue Growth Rate: 16.53\%, Gross Profit Margin: 68.84\%, Operating Profit Margin: 68.90\%, Net Profit Margin: 51.25\%. \\
    \\
    \textbf{[Instruction]}:\\
    Perform a comprehensive analysis of the company's income statement, focusing on the following aspects:

1. Revenue Analysis: Provide an overview of total revenue, including Year-over-Year (YoY) and Quarter-over-Quarter (QoQ) comparisons. Break down revenue sources to identify key contributors and trends.

2. Cost Analysis: Examine the Cost of Goods Sold (COGS) to identify potential cost control issues.

3. Profitability Metrics: Review gross margin, operating margin, and net profit margin to evaluate cost efficiency, operational effectiveness, and overall profitability.

4. Investor Perspective: Analyze Earnings Per Share (EPS) to gauge investor sentiment.

5. Benchmark Comparison: Compare these metrics with historical data to identify growth patterns, profitability trends, and operational challenges.

Summarize your findings in a single paragraph under 200 words.\\
    \bottomrule
    \end{tabular}
    \caption{Income Extraction Agent prompt example.}
    \label{tab:income_extraction_prompt}
\end{table*}

\begin{table*}[h]
    \centering

    \begin{tabular}{p{15cm}}
    \toprule
    \textbf{[Analysis Date]}:
    2024-11-05 \\
    \\
    \textbf{[Company Name]}:
    Kweichow Moutai Co., Ltd.\\
    \\ 
    \textbf{[Balance Sheet]}:\\
    Date: 2024-09-30, Total Current Assets: 238,419,577,361.93, Total Non-Current Assets: 48,351,145,95.76, Cash and Cash Equivalents: 60,084,745,614.27, Accounts Receivable: 10,131,702.74, Inventory: 48,224,880,009.62, Total Current Liabilities: 38,723,890,549.12, Total Non-Current Liabilities: 358,668,170.4, Accounts Payable: 2,652,185,045.16, Undistributed Profit: 192,903,581,645.13, Total Liabilities: 39,082,558,719.52, Total Owner's Equity: 247,688,164,238.17, Total Assets: 286,770,722,957.69, Total Liabilities and Owner's Equity: 286,770,722,957.69;\\
    Date: June 30, 2024, Total Current Assets: 232,257,040,097.43, Total Non-Current Assets: 46,950,221,069.20, Cash and Cash Equivalents: 56,840,349,530.82, Accounts Receivable: 1,358,595.12, Inventory: 47,766,478,732.37, Net Value of Fixed Assets: 19,833,577,460.59, Total Current Liabilities: 50,806,881,664.47, Total Non-Current Liabilities: 355,569,852.88, Accounts Payable: 3,291,376,926.17, Undistributed Profits: 173,771,640,509.99002, Total Liabilities: 51,162,451,517.35, Total Owner's Equity: 228,044,809,649.28, Total Assets: 279,207,261,166.63, Total Liabilities and Owner's Equity: 279,207,261,166.63;\\
    ...\\
    Date: 2022-12-31, Total Current Assets: 216,611,435,672.92, Total Non-Current Assets: 37,889,390,423.10, Cash and Cash Equivalents: 58,274,318,733.23, Accounts Receivable: 20,937,144.00, Inventory: 38,824,374,236.24, Net Value of Fixed Assets: 19,743,689,764.66, Total Current Liabilities: 49,065,668,798.38, Total Non-Current Liabilities: 497,076,033.78, Accounts Payable: 2,408,371,053.69, Undistributed Profits: 161,278,002,807.10, Total Liabilities: 49,562,744,832.16, Total Owners' Equity: 204,938,081,263.86, Total Assets: 254,500,826,096.02, Total Liabilities and Owners' Equity: 254,500,826,096.02. \\
    \\
    \textbf{[Instruction]}:
    Conduct a detailed analysis of the company's balance sheet, focusing on the following areas:

1. Asset Structure: Examine the composition of assets to assess financial stability and operational efficiency.

2. Liabilities and Shareholders' Equity: Evaluate liquidity by comparing current assets to current liabilities and assess solvency through long-term debt ratios. Analyze shareholders' equity to determine long-term investment potential.

3. Historical Comparison: Compare these metrics with data from previous years to identify financial trends and signs of improvement or deterioration.

4. Strategic Assessment: Provide a strategic overview of the company’s financial leverage, asset management, and capital structure.

Summarize your findings in a paragraph of no more than 150 words, highlighting key insights into the company's financial health and future prospects.\\
    \bottomrule
    \end{tabular}
        \caption{Balance Extraction Agent: prompt example.}
    \label{tab:balance_extraction_prompt}
\end{table*}

\begin{table*}[h]
    \centering

    \begin{tabular}{p{15cm}}
    \toprule
    \textbf{[Analysis Date]}:
    2024-11-05 \\
    \\
    \textbf{[Company Name]}:
    Kweichow Moutai Co., Ltd.\\
    \\ 
    \textbf{[Cash Flow Statement]}:\\
    Date: 2024-09-30, Net Cash Flow from Operating Activities: 44,421,386,217.45, Net Cash Flow from Investing Activities: -605,409,055.95, Net Cash Flow from Financing Activities: -38,989,316,464.81, Net Increase in Cash and Cash Equivalents: 4,826,595,257.6, Cash Received from Investment Recovery: 6,200,000,000.0, Cash Received from Investment Income: 68,399,488.96, Cash Paid for Acquisition of Fixed Assets, Intangible Assets, and Other Long-term Assets: 2,874,366,279.39,  Cash Paid for Distribution of Dividends, Profits, or Interest Payments: 38,942,461,843.34;\\
    Date: 2024-06-30, Net Cash Flow from Operating Activities: 36,621,833,812.63, Net Cash Flow from Investing Activities: -2,898,860,221.14, Net Cash Flow from Financing Activities: -38,815,639,978.07, Net Increase in Cash and Cash Equivalents: -5,093,365,381.42, Cash Received from Investment Recovery: 2,600,000,000.0, Cash Received from Investment Income: 31,308,120.34, Cash Paid for Acquisition of Fixed Assets, Intangible Assets, and Other Long-term Assets: 1,530,991,618.84,  Cash Paid for Distribution of Dividends, Profits, or Interest Payments: 38,786,363,272.8;\\
    ...\\
    Date: 2022-12-31, Net cash flow from operating activities: 36,698,595,830.03, Net cash flow from investing activities: -5,536,826,334.90, Net cash flow from financing activities: -57,424,528,979.83, Net increase in cash and cash equivalents: -26,261,848,396.69, Cash received from investment income: 5,880,000.0, Cash paid for the acquisition of fixed assets, intangible assets, and other long-term assets: 5,306,546,416.54, Cash paid for dividends, profit distributions, or interest payments: 57,370,196,191.46.\\
    \\
    \textbf{[Instruction]}:\\
    Based on the provided cash flow statement data, comprehensively evaluate the company's cash flow situation. Focus on the following aspects:

1. Operating Activities: Use ``Net Cash Flow from Operating Activities'' to measure the profitability of core business operations.

2. Investing Activities: Review the ``Net Cash Flow from Investing Activities'' and the specific cash flows related to investments and capital expenditures, such as ``Cash Received from Investment Recovery'' and ``Cash Paid for the Acquisition of Fixed Assets, Intangible Assets, and Other Long-term Assets''.

3. Financing Activities: Assess the ``Net Cash Flow from Financing Activities'' and analyze changes in debt and equity, such as ``Cash Paid for Debt Repayment'' and ``Cash Paid for Dividends, Profits Distribution, or Interest Payments''.

Finally, compare the current period data with historical data to identify trends, assess sustainability and liquidity risks, and briefly analyze the company's cash management efficiency, liquidity status, and future growth potential or financial challenges. Limit to 150 words.\\
    \bottomrule
    \end{tabular}
        \caption{Cash Extraction Agent prompt example.}
    \label{tab:cash_extraction_prompt}
\end{table*}

\begin{table*}[h]
    \centering

    \begin{tabular}{p{15cm}}
    \toprule
    \textbf{[Analysis Date]}:
    2024-11-05 \\
    \\
    \textbf{[Company Name]}:
    Kweichow Moutai Co., Ltd.\\
    \\ 
    \textbf{[Income Extraction Agent Output]}:\\
    In the third quarter of 2024, Guizhou Moutai achieved significant revenue growth, with a year-on-year increase of 16.91\% and a quarter-on-quarter increase of 47.54\%, indicating strong market demand. The cost of sales was well controlled, with a gross profit margin of 68.18\%, slightly lower than in previous quarters but still demonstrating high cost efficiency. The operating profit margin and net profit margin were 68.22\% and 51.19\%, respectively, showing a slight decline, which may reflect a slight increase in operating costs. Basic earnings per share were 48.42 yuan, indicating that the company's profitability remains strong, and investor confidence is likely to remain stable. Compared to historical data, although the gross profit margin and net profit margin slightly decreased, the overall profitability trend remains positive, demonstrating continued growth potential and market advantage. \\
    \\
    \textbf{[Balance Extraction Agent Output]}: \\
    1. Asset Structure: The proportion of current assets remains stable, primarily composed of cash and inventories, reflecting strong short-term debt repayment capability and effective inventory management. ... 4. Strategic Evaluation: Kweichow Moutai operates primarily on self-owned capital, maintaining a low leverage ratio and high asset management efficiency. The company's capital structure is conservative yet highly stable. Kweichow Moutai exhibits strong financial stability, excellent asset liquidity, extremely low debt risk, substantial shareholder return potential, and promising long-term development prospects. \\
    \\
    \textbf{[Cash Extraction Agent Output]}:\\
    1. Operating Activities: Continued to be strong in 2024, with net cash flow from operating activities reaching 100.231 billion yuan in the first three quarters, showing a significant increase compared to the same period in 2023. The profitability of core business remained stable, indicating strong market demand and operational efficiency. ... 3. Financing Activities: The net cash flow from financing activities was significantly negative, primarily due to dividend payments. There were no borrowings or equity financing, with funds relying on operating activities. While the dividend policy attracts shareholders, it limits cash reserves. There was an overall abundance of cash and cash equivalents, with a net increase of 4.827 billion yuan in Q3 2024. The core business is robust, but high dividends and investment expenditures might impact long-term liquidity. A balance must be struck between expansion and shareholder return strategies.\\
\\
    \textbf{[Instruction]}: \\
    Based on the provided financial data for the stock, summarize its financial performance and analyze its recent trends, focusing on key indicators such as revenue, profit, liabilities, and cash flow. Generate a single paragraph for inclusion in an equity research report, not exceeding 200 words, and ensure the inclusion of specific financial figures. Please return the text in the following JSON format: \texttt{
  \{"paragraph": "Content of a single paragraph",
  "title": "Concise and appropriate title generated based on the paragraph content"\} }\\
    \bottomrule
    \end{tabular}
        \caption{Finance Analysis Agent prompt example.}
    \label{tab:finance_ana_prompt}
\end{table*}

\begin{table*}[h]
    \centering

    \begin{tabular}{p{15cm}}
    \toprule
    \textbf{[Analysis Date]}:
    2024-11-05 \\
    \\
    \textbf{[Company Name]}:
    Kweichow Moutai Co., Ltd.\\
    \\ 
    \textbf{[Core News]}:\\
    News Date: 2024-10-29. News Summary: Kweichow Moutai Group plans to collaborate with He Eye Care to establish the Zhaohua Health Fund, with a total investment of 797 million RMB. Of this, Moutai Zhaohua will invest 554 million RMB, accounting for 69.511\%. The fund will focus on new-generation healthcare, nutritional food therapy, and synthetic biology, aiming to use eye health as an entry point. Previously, Moutai had invested in synthetic biology company Hongmo Bio, highlighting its strategic positioning in the bioeconomy and emerging industries. Moutai Chairman Ding Xiongjun emphasized that bioeconomy, centered around synthetic biology, is a key direction for future development. \\
    News Date: 2024-10-28. News Content: The 111th National Sugar and Wine Expo opened on October 29 at the Shenzhen International Convention and Exhibition Center, attracting 4,000 domestic and international food and beverage companies. Kweichow Moutai's booth, designed with the classic ``red on top, white on the bottom'' color scheme, prominently displayed its products and culture, emphasizing its shift from ``selling liquor'' to ``selling lifestyle''. Moutai attracted a large number of visitors, especially younger consumers, through rich interactive experiences and diverse consumption scenarios. The company also showcased its efforts in integrating liquor and tourism and in cultural dissemination, further expanding its market boundaries and enhancing its brand influence. \\
    ... \\
    News Date: 2024-10-18. News Content: Recently, Guo Jiang 1935 was accused of imitating Kweichow Moutai, raising concerns about market irregularities in the liquor sector. Kweichow Moutai's Moutai 1935 achieved over 10 billion RMB in sales in a short period, but many counterfeits have appeared in the market. Guo Jiang 1935 has a similar appearance to Moutai 1935, leading to consumer confusion. This imitation behavior not only misguides consumers but also potentially infringes on intellectual property rights. Additionally, the price fluctuations of Moutai 1935 are influenced by e-commerce promotions and counterfeits. Consumers are advised to purchase from legitimate channels and carefully distinguish between brands and packaging. Liquor companies need to strengthen self-regulation, improve product quality, while regulatory authorities should increase market supervision to maintain order and protect consumer rights. \\
    \\
    \textbf{[Instruction]}:\\
    Please analyze the market impact on the given stock based on the core news provided. Discuss potential short-term and long-term trends, considering market sentiment and the influence of external factors. Generate a single paragraph for inclusion in an equity research report, not exceeding 200 words. Please return the text in the following JSON format: \texttt{\{``paragraph'': ``Content of a single paragraph'', ``title'': ``A concise and appropriate title generated based on the paragraph content''
    \}} \\
    \bottomrule
    \end{tabular}
        \caption{News Analysis Agent prompt example.}
    \label{tab:news_ana_prompt}
\end{table*}

\begin{table*}[h]
    \centering

    \begin{tabular}{p{15cm}}
    \toprule
    \textbf{[Analysis Date]}:
    2024-11-05 \\
    \\
    \textbf{[Company Name]}:
    Kweichow Moutai Co., Ltd.\\
    \\ 
    \textbf{[Company Announcements]}: \\
     1. Description of the Company's Industry and Main Business: The company’s primary business is the production and sales of Moutai liquor and its series products. Its flagship product, ``Guizhou Moutai'' is recognized as the pioneer and quintessential representative of China’s Maotai-flavor baijiu. It is a blend of national geographical indication products, organic food, and intangible cultural heritage. The company’s marketing network spans the domestic market and 64 countries and regions across five continents. For years, the company has upheld the pursuit of product quality, carefully maintained its brewing ecosystem, innovatively preserved traditional brewing methods, and continuously driven high-quality development and modernization efforts. The company operates on the following business model: procurement of raw materials, production of products, and sales of products. This year, China’s economy has been generally stable, continuing its upward trend with domestic demand recovering steadily, external demand improving, and high-quality development advancing solidly. As policies to expand domestic demand intensify, the focus of economic policies has shifted more towards improving people’s livelihoods and promoting consumption, driving further expansion and upgrading of consumption. 
     
     2. Analysis of Core Competitiveness: The company boasts ``Five Core Competitiveness'' factors-quality, brand, environment, craftsmanship, and culture-alongside “Four Core Momentum” elements, namely: Unique geographic origin protection, Irreplicable microbial community, Centuries-old unique brewing craftsmanship, Long-term base liquor storage resources. 
     
     3. Discussion and Analysis of Operating Conditions: This year, guided by Xi Jinping Thought on Socialism with Chinese Characteristics for a New Era, the company has earnestly implemented the spirit of the 20th National Congress of the CPC, the Central Economic Work Conference, and President Xi’s important remarks during his inspection of Guizhou. Adhering to the belief that “quality is the soul of life,” the company has focused on its core liquor business, continuously strengthening the drivers of reform, openness, and innovation. Its overall trajectory is steady and positive, as reflected in the following three aspects: ... \\
    \\
    \textbf{[Instruction]}: \\
    Based on the provided company announcements, please explore the company's strategic direction and future development potential, and assess its competitive position within the industry. Generate a single paragraph for an equity research report, not exceeding 200 words.\\
    \bottomrule
    \end{tabular}
    \caption{Status Analysis Agent prompt example.}
    \label{tab:status_ana_prompt}
\end{table*}

\begin{table*}[h]
    \centering

    \begin{tabular}{p{15cm}}
    \toprule
    \textbf{[Analysis Date]}:
    2024-11-05 \\
    \\
    \textbf{[Company Name]}:
    Kweichow Moutai Co., Ltd.\\
    \\ 
    \textbf{[Finance Analysis Agent Output]}:\\
    Kweichow Moutai's Strong Financial Performance in Q3 2024 with Robust Liquidity: Guizhou Moutai demonstrated robust market performance in the third quarter of 2024, with revenue growing by 16.91\% year-on-year and 47.54\% quarter-on-quarter, despite a slight decline in gross profit margin to 68.18\%. Operating and net profit margins stood at 68.22\% and 51.19\%, respectively, reflecting a marginal increase in operating costs. The company's balance sheet reveals current assets significantly exceeding current liabilities, with a current ratio exceeding 4 and virtually no long-term debt, indicating extremely low financial risk. The net cash flow from operating activities reached RMB 100.231 billion, underscoring strong market demand and operational efficiency. Although net cash flow from investing activities was negative, reflecting expansion, cash and cash equivalents still increased by RMB 4.827 billion, indicating ample liquidity. Under its high dividend policy, Guizhou Moutai must carefully balance business expansion and shareholder returns. \\
    \\
    \textbf{[News Analysis Agent Output]}:\\
    Strategic Layout and Brand Influence Supporting Long-term Stable Development: Recently, Kweichow Moutai has demonstrated positive market dynamics in several areas. The company’s collaboration with He Eye Specialist Hospital to establish a health fund highlights its strategic layout in the bio-economy sector, which may drive long-term growth. Moutai received an upgrade in its MSCI ESG rating, reflecting its efforts in sustainable development. Amid the differentiation trend in the liquor industry, Moutai continues to attract investor attention with its robust performance growth and high dividend policy. The reduction of spirits taxes in Hong Kong could accelerate Moutai's internationalization process, though cultural barriers in overseas markets still warrant attention. In the short term, market sentiment may fluctuate due to policy changes and counterfeit product issues, but in the long run, Moutai's brand influence and strategic layout will underpin its steady development.\\
    \\
    \textbf{[Status Analysis Agent Output]}:\\
    Solid Growth and Brand Value Enhancement: Guizhou Moutai demonstrated strong growth momentum in the first half of 2024, achieving double-digit increases in both revenue and net profit, underscoring its solid position in the high-end liquor market. The company further enhanced its market competitiveness by strengthening brand value and advancing digital transformation. Moutai's ``five core competencies'' and ``four core momentum drivers'' enable it to maintain a leading position in the industry. Looking ahead, with domestic consumption upgrades and international market expansion, Moutai is expected to sustain its growth trajectory and solidify its leadership among global liquor brands.\\
    \\
    \textbf{[Instruction]}:\\
    Please refer to the above agents' output results and analyze the potential risk factors for this stock. Provide at least three distinct risks, each summarized in no more than 10 words. Please convert your output format to JSON, which can be directly loaded by the json.loads() function: \texttt{\{``risks'': [``risk1'', ``risk2'', ..., ``riskn'']\}}\\
    \bottomrule
    \end{tabular}
    \caption{Risks Analysis Agent prompt example.}
    \label{tab:risks_ana_prompt}
\end{table*}

\begin{table*}[h]
    \centering

    \begin{tabular}{p{15cm}}
    \toprule
    \textbf{[Analysis Date]}:
    2024-11-05 \\
    \\
    \textbf{[Company Name]}:
    Kweichow Moutai Co., Ltd.\\
    \\ 
     \textbf{[Finance Analysis Agent Output]}:\\
    Kweichow Moutai's Strong Financial Performance in Q3 2024 with Robust Liquidity: Guizhou Moutai demonstrated robust market performance in the third quarter of 2024, with revenue growing by 16.91\% year-on-year and 47.54\% quarter-on-quarter, despite a slight decline in gross profit margin to 68.18\%. Operating and net profit margins stood at 68.22\% and 51.19\%, respectively, reflecting a marginal increase in operating costs. The company's balance sheet reveals current assets significantly exceeding current liabilities, with a current ratio exceeding 4 and virtually no long-term debt, indicating extremely low financial risk. ... \\
    \\
    \textbf{[News Analysis Agent Output]}:\\
    Strategic Layout and Brand Influence Supporting Long-term Stable Development: Recently, Kweichow Moutai has demonstrated positive market dynamics in several areas. The company’s collaboration with He Eye Specialist Hospital to establish a health fund highlights its strategic layout in the bio-economy sector, which may drive long-term growth. Moutai received an upgrade in its MSCI ESG rating, reflecting its efforts in sustainable development. Amid the differentiation trend in the liquor industry, Moutai continues to attract investor attention with its robust performance growth and high dividend policy. ...\\
    \\
    \textbf{[Status Analysis Agent Output]}:\\
    Solid Growth and Brand Value Enhancement: Kweichow Moutai demonstrated strong growth momentum in the first half of 2024, achieving double-digit increases in both revenue and net profit, underscoring its solid position in the high-end liquor market. The company further enhanced its market competitiveness by strengthening brand value and advancing digital transformation. Moutai's ``five core competencies'' and ``four core momentum drivers'' enable it to maintain a leading position in the industry. ...\\
    \\
    \textbf{[Risks Analysis Agent Output]}: \\
    Public opinion risks, Policy fluctuation risks, Counterfeit product risks, Market cultural barriers.\\
    \\
    \textbf{[History Market Indices]}:\\
     2024-10-08: 4256.1, 2024-10-09: 3955.98, 2024-10-10: 3997.79, ... , 2024-11-01: 3890.02, 2024-11-04: 3944.76, 2024-11-05: 4044.57
    \\
    \textbf{[Historical Stock Prices]}:\\
    2024-10-08: 1723.0, 2024-10-09: 1595.15, 2024-10-10: 1640.0, ... , 2024-11-01: 1533.75, 2024-11-04: 1548.2, 2024-11-05: 1576.99\\
    \\
    \textbf{[Instruction]}:\\
    Based on the above information, please provide investment advice and predict the trend of the company's stock over the next three weeks. If the expected increase in the company's stock exceeds the rise of the CSI 300 Index, assign a "Buy" rating; if it is lower, assign a "Sell" rating. Write an equity research report not exceeding 200 words. Please return a text in the following JSON format: \texttt{\{``paragraph'': ``Content of a single paragraph'', ``title'': ``A concise and appropriate title generated from the paragraph content'', ``rating'': ``Buy/Sell''\}} \\
    \bottomrule
    \end{tabular}
        \caption{Prediction Agent prompt example.}
    \label{tab:prediction_prompt}
\end{table*}

\begin{table*}[h]
    \centering

    \begin{tabular}{p{15cm}}
    \toprule
    \textbf{[Expert-written Equity Research Reports]}: \\
    \texttt{Content of reports 1.}\\
    \texttt{Content of reports 2.}\\
    ...\\
    \texttt{Content of reports n.}\\
    \\
    \textbf{[Generated Equity Research Report]}:\\
    \texttt{Content of generated Equity Research Report.}\\
    \\
    \textbf{[Instruction]}:\\
    Please refer to the above Export-written Equity Research Reports and make appropriate modifications and corrections to the Generated Equity Research Report. Please note the following:

1. Maintain the original JSON format and structure.

2. Improve the accuracy and logic of the report content.

3. Ensure consistency of all data and analytical results.

4. Correct any grammatical errors or typos.

5. Ensure the correct use of industry terminology.

Do not alter the stock ratings. Only make appropriate modifications and corrections to the paragraph content without altering the original stock ratings of the report. Directly return the modified original JSON formatted report without explaining the reasons.\\
    \bottomrule
    \end{tabular}
        \caption{Prompt used in Expert-written ERRs Corrector.}
    \label{tab:corrector_prompt}

\end{table*}

\end{document}